\documentclass{article} % For LaTeX2e
\usepackage{iclr2026_conference,times}

\usepackage{graphicx} % Required for inserting images
\usepackage{amsmath}
\usepackage{amssymb}
\usepackage{amsthm}

\theoremstyle{plain} % it：theorem, lemma, proposition, corollary
\newtheorem{theorem}{Theorem}
\newtheorem{lemma}[theorem]{Lemma}

\newtheorem{corollary}[theorem]{Corollary}

\usepackage{booktabs}
\usepackage{multirow}
\usepackage[table,xcdraw]{xcolor}
\usepackage[labelfont=bf]{caption}
\usepackage{wrapfig}

\usepackage{hyperref}
\usepackage{url}

\usepackage[ruled,vlined]{algorithm2e}

\usepackage{wrapfig}

\usepackage{ifthen}
\usepackage{color}
\newcommand{\ShowComments}{no}

\ifthenelse{\equal{\ShowComments}{yes}}{

\newcommand{\YC}[1]{{\color{brown}[YC: #1]}}
\newcommand{\PY}[1]{{\color{blue}PY: #1}}
\newcommand{\PYB}[1]{{\color{blue}[PY: #1]}}
}{
\newcommand{\YC}[1]{}
\newcommand{\PY}[1]{}
\newcommand{\PYB}[1]{}
}
\usepackage{enumitem}
\newcommand{\TBD}[1]{{\color{brown}[TBD]}}

\title{CarBoN: Calibrated Best-of-N Sampling\\ Improves Test-time Reasoning}

\iclrfinalcopy

\author{
Yung-Chen Tang$^{1,2}$,
Pin-Yu Chen$^{3}$,
Andrea Cavallaro$^{1,2}$ \\
$^{1}$EPFL \quad
$^{2}$Idiap Research Institute \quad
$^{3}$IBM Research \\
}

\begin{document}

\maketitle

\begin{abstract}

Allocating more computation during inference time (test-time scaling) improves language model performance, especially for reasoning tasks.
However, popular methods like Best-of-$N$ sampling often show diminishing returns as $N$ increases.
To address this inefficiency, we introduce a general \textbf{test-time calibration framework} that adaptively modifies the model toward high-reward reasoning paths, with theoretical guarantees of improving the lower bound of expected reward under finite sampling, all without large language model (LLM) retraining.
Within this framework, we propose \textbf{CarBoN} (Calibrated Best-of-$N$), a two-phase method that first explores the solution space and then learns a calibration of the logits via an input-specific temperature $T$ and additive shift vector $\delta$, guiding generation toward more reliable reasoning.
Experiments on MATH-500 and AIME-2024 show that CarBoN improves efficiency, with up to $4\times$ fewer rollouts to reach the same accuracy, while often achieving higher accuracy under fixed budgets.
We also analyze the complementary roles of $T$ and $\delta$ in balancing output diversity and correctness, and demonstrate that the framework also generalizes to step-level sampling strategies such as beam search.
For more information, please refer to our project page at 
\href{https://huggingface.co/spaces/TrustSafeAI/Test-Time-Calibration}{huggingface.co/spaces/TrustSafeAI/Test-Time-Calibration}.

\end{abstract}
\section{Introduction}

Test-time scaling (TTS) is a practical alternative to ever-larger training, enabling models to “think longer” at inference by allocating additional computation to reasoning.
Methods such as chain of thought \citep{openai2024learning, guo2025deepseek}, sequential reasoning \citep{wang2022self, qu2024recursive, shinn2023reflexion}, and parallel sampling \citep{snell2024scaling, beeching2024scalingtesttimecompute, puri2025probabilistic, liu2025can} demonstrate that increased test-time effort consistently improves performance without retraining.
% A key advantage of TTS is that smaller LLMs match or even outperform larger models, offering a more cost-efficient and flexible way to adapt inference strategies to different tasks.
As these studies suggest, TTS allows smaller LLMs to match or even outperform larger ones, providing a more cost-efficient and flexible inference strategy.

Despite these benefits, simply increasing test-time compute does not guarantee optimal performance.
Recent work has shown that inference without effective verification is often sub-optimal, as models may spend additional computation on low-quality reasoning paths \citep{setlur2025scaling}.
To overcome this inefficiency, we propose a general \textbf{test-time calibration framework} that strategically reallocates the inference budget by leveraging feedback from a verifier or reward model during inference.
Rather than treating generation as a fixed forward pass, the model adaptively steers toward high-reward (likely correct) regions, improving reasoning reliability under a fixed query budget.

\begin{figure*}[t!hbp]
    \centering
    \includegraphics[width=0.95\linewidth]{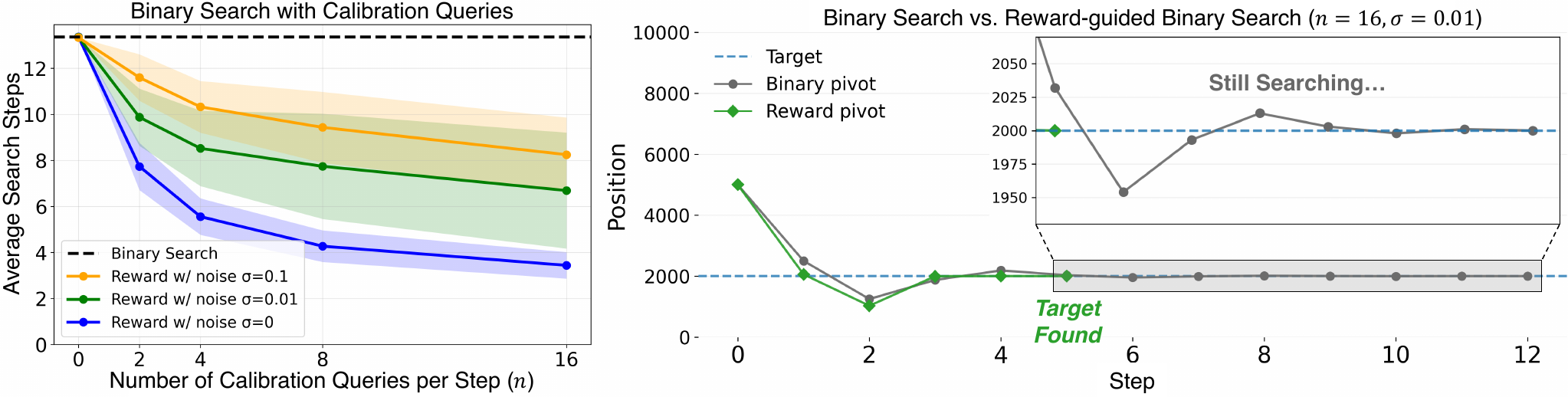}
    \vspace{-0.4em}
    \caption{\textbf{Reward‑guided calibration accelerates binary search.}
    Left: Increasing per‑step noisy reward (inverse‑distance signal + noise) lowers average search steps versus vanilla.
    Right: Example showing reward guidance converges early; vanilla keeps oscillating. See Appendix~\ref{appendix:toy-example} for details. %\PYB{mention where to find details}\YC{addressed}
    }
    \label{fig:toy_example}
\end{figure*}

\paragraph{Why calibration for TTS? A motivating example of reward-based binary search.}
% To illustrate the benefit of calibration, we view LLM inference with increased test-time compute as a search over candidate solutions, similar to binary search for a hidden target.
% In our toy example, binary search is augmented with noisy reward guidance: at each step, $k$ points in the bracket are probed for reward, given by noisy inverse distance to the target.
% Figure~\ref{fig:toy_example} shows that more per-step calibration queries consistently reduce search steps, even with noise. \PYB{How fast? Make it more concrete}
% This demonstrates that reward feedback accelerates convergence \PYB{because it changes the sampling distribution} and motivates our use of reward signals for test-time calibration. \PYB{Need to emphasise the baseline and naive TTS (calibration number = 0)} 
% Let the task be finding a target in $[0,10^4]$. 
% Calibration means that before each search step the model can query $n$ candidate points for reward, which is the inverse distance to the target plus noise. The baseline binary search, corresponding to naive TTS ($n=0$) \PYB{we never defined $n$}, requires $13.3$ steps on average. Increasing $n$ significantly accelerates convergence: for example, with $n=16$ the search depth is reduced by up to $74\%$ (see Figure~\ref{fig:toy_example}, left).
% Figure~\ref{fig:toy_example} (right) shows an example run where calibration quickly converges to the target, while vanilla binary search continues oscillating. This example highlights that reward feedback \PY{for calibration} reshapes the sampling distribution and motivates its use for TTS.
Let the task be finding a target in $[0,10^4]$. 
Calibration means that before each search step the model can query $n$ candidate points for reward, where $n$ denotes the number of reward queries per step.
The reward is defined as the inverse distance to the target plus noise.
The baseline binary search, corresponding to naive TTS ($n=0$), requires $13.3$ steps on average. Increasing $n$ significantly accelerates convergence: for example, with $n=16$ the search depth is reduced by up to $74\%$ (see Figure~\ref{fig:toy_example}, left).
Figure~\ref{fig:toy_example} (right) shows an example run where calibration quickly converges to the target, while vanilla binary search continues oscillating.
This example highlights that reward feedback for calibration reshapes the sampling distribution and motivates its use for TTS.
\PYB{we never defined $n$} \YC{added.}

%\PYB{How fast? Make it more concrete}
%\PYB{because it changes the sampling distribution}
%\PYB{Need to emphasize the baseline and naive TTS %(calibration number = 0)}
%\YC{addressed}

% Building on this principle, our framework reuses sampled completions that are normally discarded in parallel sampling methods to extract reward signals and perform calibration.
% Within this framework, we introduce \textbf{CarBoN} (Calibrated Best-of-$N$), which allocates part of the budget to exploration and calibration, then focuses the remaining budget on high-reward regions using logit calibration.
% Reusing high-scoring answer selected by reward model enhances answer quality and query efficiency without retraining, under the same inference budget.

Building on this principle, our framework reuses sampled completions that are normally discarded in parallel sampling methods \citep{wang2022self,brown2024large,snell2024scaling} to extract reward signals and perform calibration.
Within this framework, we introduce \textbf{CarBoN} (Calibrated Best-of-$N$). Without modifying the original LLM, our framework allocates part of the budget to exploration and calibration, then focuses the remaining budget on high-scoring regions using logit calibration.
Reusing high-scoring answer selected by reward model enhances answer quality and query efficiency, under the same inference budget.

\begin{figure*}[th]
    \centering
    \includegraphics[width=\linewidth]{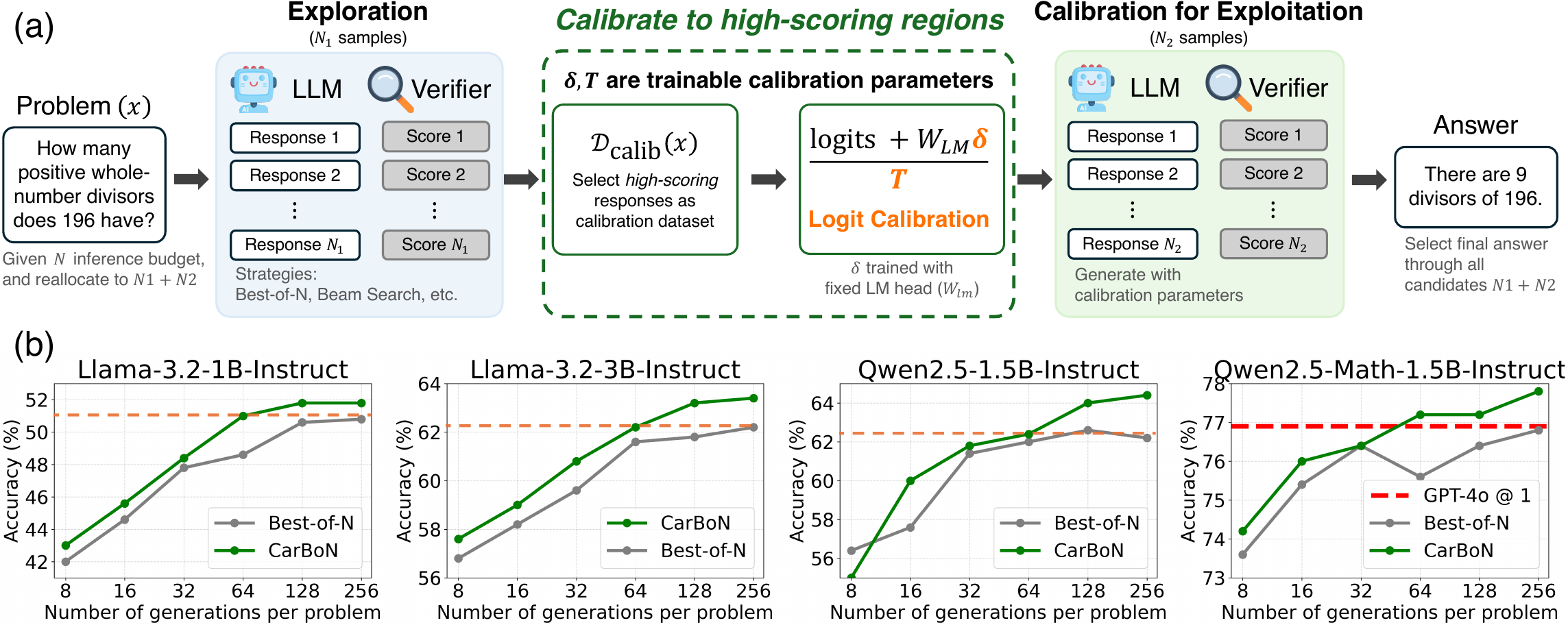}
    \vspace{-0.8em}
    \caption{\textbf{(a) Test-time calibration framework.} With a rollout budget $N = N_1 + N_2$, the model first explores by generating and scoring $N_1$ candidate responses. The model then learns calibration parameters $(\delta, T)$ from high-scoring responses, , using them to adjust the logits for the remaining $N_2$ generations. The final answer is selected from all $N$ candidates.
    \textbf{(b) MATH-500 Results.} CarBoN improves weighted Best-of-N accuracy across four models. For all models, calibrated accuracy at $N=64$ (orange dash line) matches or exceeds uncalibrated accuracy at $N=256$, corresponding to up to a $4\times$ reduction in rollout budgets. Notably, with Qwen2.5-Math-1.5B-Instruct at $N=64$, CarBoN surpasses GPT-4o (red dashed line), while uncalibrated Best-of-N with $N=256$ does not.}
    \vspace{-0.8em}
    \label{fig:main}
\end{figure*}

\noindent
Our main contributions are summarized as follows:
\begin{itemize}[leftmargin=*]
    \item \textbf{Test-Time Calibration Framework.} We introduce a test-time calibration framework that reallocates the inference budget (Figure~\ref{fig:main}a). 
    Applied to Best-of-$N$, CarBoN first explores diverse candidates to identify high-scoring regions, then uses logit calibration to focus the remaining budget on high-scoring areas, improving accuracy under fixed rollout budget.
    
    \item \textbf{Theoretical Guarantees.} 
    We provide formal proofs showing that optimal calibration parameters exist, which improve the expected reward's lower bound under finite sampling and strictly outperform the uncalibrated baseline. 
    
    \item \textbf{CarBoN Empirically Improves Test-Time Reasoning.} Across multiple models and benchmarks, including MATH-500 and the more challenging AIME-2024, CarBoN achieves higher or comparable accuracy with fewer queries than uncalibrated models, showing benefits for both general-purpose and math-specialized models. %\PYB{how about AIME?} \YC{added}

    \item \textbf{Calibration Insights and Generalization.} In CarBoN, we find that temperature ($T$) controls output distribution sharpness, delta ($\delta$) corrects token-level biases, and together they balance diversity and correctness to improve test-time reasoning. We further generalize test-time calibration beyond Best-of-$N$, applying to step-level sampling (beam search) to demonstrate broader applicability.
\end{itemize}

% At $N=256$, calibration further improves accuracy by over $1\%$.

% Furthermore, parallel sampling often discards most generated completions after scoring, wasting computation and limiting efficiency. Our framework reuses sampled completions for calibration, maximizing query utility and answer quality.
% Within this framework, we present \textbf{CarBoN} (Calibrated Best-of-N) as a simple yet effective instantiation.
% CarBoN uses part of the inference budget for exploration and calibration, then focuses the rest on high-reward regions via lightweight logit calibration. This approach enables better utilization of a fixed query budget and leads to improved reasoning and performance.
\section{Related Work}

\vspace{-0.8em}
\paragraph{Reasoning with Intermediate Steps.}
% Recent advances in large language models have focused on enhancing their reasoning abilities by encouraging models to generate detailed intermediate steps before producing final answers.
% Techniques such as chain-of-thought prompting \citep{wei2022chain, kojima2022large}, least-to-most prompting \citep{zhou2022least}, and learned reasoning policies \citep{yue2023mammoth, yu2023metamath, wang2023math, openai2024learning, anthropic2025tracing, guo2025deepseek} enable models to improve complex problem solving by “thinking longer” during inference, effectively utilizing a large output token budget to perform multi-step reasoning within a single forward pass.
% From the perspective of test-time compute, these approaches scale inference primarily by allocating more output tokens to reasoning.
%
% However, these approaches often face practical limitations related to the model’s context window and KV cache size, as longer chains of thought consume more memory and computational resources during inference \citep{brown2020language}.
% This restricts the feasible length of reasoning and may lead to diminishing returns when scaling inference-time compute purely by increasing token generation length.

Recent work has improved LLM reasoning by encouraging generation of intermediate steps, e.g., chain-of-thought \citep{wei2022chain, kojima2022large}, least-to-most prompting \citep{zhou2022least}, and learned reasoning policies \citep{yue2023mammoth, yu2023metamath, wang2023math, openai2024learning, anthropic2025tracing, guo2025deepseek}.
These methods use a large token budget for \PYB{remove to?} \YC{addressed.} multi-step reasoning within a single forward pass, effectively “thinking longer” at inference.
However, they remain limited by context window and KV cache constraints, which can restrict feasible reasoning length and make naive scaling of token generation inefficient.

\vspace{-0.8em}
\paragraph{Iterative Refinement.}
Sequential revision methods improve outputs by feeding previous answers back. Recursive reasoning \citep{qu2024recursive} uses multiple critique rounds to correct mistakes; the authors note early errors can propagate and gains often diminish after a few iterations. Reflective prompting \citep{shinn2023reflexion} adds self-assessment but its effectiveness is limited by memory and reflection quality. Overall, these methods enhance accuracy without retraining but increase latency, RAM usage, and computation linearly, and repeated refinement may yield diminishing returns.

\vspace{-0.8em}
\paragraph{Parallel Sampling Strategies.}
Parallel sampling methods can be divided into two groups.
The first generates complete candidate answers per query without intermediate evaluation.
This includes self-consistency (majority voting) and best-of-$N$ (BoN), where the former selects the most frequent answer and the latter scores each candidate with a reward model, choosing the highest-scoring answer \citep{wang2022self, brown2024large}.
Best-of-$N$ generally outperforms majority voting, as verifier-free selection is suboptimal \citep{setlur2025scaling}.
This approach is simple, efficient, and provides a practical baseline.

The second group consists of step-level methods, which evaluate candidates at each generation step for finer-grained control and typically higher-quality results.
These include Beam Search \citep{snell2024scaling}, Diverse Verifier Tree Search (DVTS) \citep{beeching2024scalingtesttimecompute}, and Particle Filtering \citep{puri2025probabilistic}, all of which are variations of Beam Search.
Beam Search maintains the top-k high-scoring beams at each step.
DVTS allows independent beams and optionally samples lower-scoring steps, balancing exploration and exploitation.
Particle Filtering converts step scores into probabilities and samples candidate steps, maintaining a diverse particle set for probabilistic inference.

Step-level methods often improve quality and diversity but are computationally costly due to repeated scoring and pruning.
In this work, we focus on Best-of-$N$ as the main baseline for test-time calibration and include Beam Search experiments to validate the step-level framework.

\PYB{If we need more space, we can briefly summarize what model calibration is using few sentences, and say we provide more background and related of calibration in Appendix}
\vspace{-0.8em}
\paragraph{Model Calibration.}
% Model calibration traditionally focuses on aligning a model’s predicted probabilities with empirical correctness, particularly in classification settings.
% Techniques such as temperature scaling \citep{guo2017calibration}, histogram binning \citep{zadrozny2001obtaining}, isotonic regression \citep{zadrozny2002transforming}, as well as more recent approaches such as Dirichlet calibration \citep{kull2019beyond} and joint input-output calibration \citep{tang2024neural}, have been widely used to improve uncertainty estimation by ensuring that confidence scores reflect true likelihoods of correctness.

% However, existing calibration methods are almost always applied in a post-hoc manner under a fixed model, where the logits are adjusted to improve the alignment between confidence and correctness.
% In our work, we borrow this notion of post-hoc adjustment under a frozen LLM, but with a different objective.
% Rather than aligning probabilities with correctness, our goal is to calibrate the model’s exploration at test time by guiding generation toward higher-reward regions of the output space, without any additional model retraining or reliance on ground-truth answers.

Model calibration traditionally aligns a model’s predicted probabilities with empirical correctness in classification, using techniques such as temperature scaling \citep{guo2017calibration}, histogram binning \citep{zadrozny2001obtaining}, isotonic regression \citep{zadrozny2002transforming}, Dirichlet calibration \citep{kull2019beyond}, and joint input-output calibration \citep{tang2024neural}.
These methods generally operate post-hoc on fixed models to improve confidence estimation.
In this work, we adopt the idea of post-hoc logits adjustment under a frozen LLM, but change the objective from correctness alignment to calibrating test-time scaling sampling, thereby shifting generation toward higher-reward regions without model retraining or relying on ground-truth labels.
\section{Test-Time Calibration}
Building on our motivation in the introduction, we observe that parallel sampling typically generates many candidate completions, of which only the highest-scoring is selected while the rest are discarded. We hypothesize that the discarded completions contain valuable signals that, if reused, can better calibrate the model’s output and improve answer quality.
This leads to our first research question (RQ):
\textbf{RQ1. \textit{How can we guide LLM inference under test-time scaling by reusing information from discarded completions in parallel sampling in order to calibrate the output distribution and enhance answer quality under a fixed compute budget?}}
\PYB{should we use TTS here?}\YC{added.}

\subsection{Balancing Exploration and Exploitation at Test-Time}
\YC{next-token-probibility}
% To address RQ1, we reframe parallel sampling as a two-phase optimization at test time: exploration and exploitation, with a total inference budget $N=N_1+N_2$. 
% Specifically, we first use $N_1$ of the inference budget to explore the output space and identify a high-scoring (high-reward) distribution. We then calibrate the model to this distribution and use the remaining budget ($N_2$) for focused sampling, thereby balancing exploration and exploitation under a fixed budget (see Figure~\ref{fig:main}).

To address RQ1, we first define logits calibration as a learnable transformation that reshapes the model’s output distribution at test time.
Formally, let $x$ be the input problem, $y=(y_1, \dots, y_T)$ a generated answer sequence, and $\theta$ the fixed LLM parameters.
The calibrated next-token distribution is then defined as:
\begin{equation}
\label{next-token-prediction}
    p_\theta(y_t \mid y_{<t}, x; \delta, T) = \operatorname{softmax}\!\left( \frac{\text{logits} + W_\mathrm{LM} \cdot \delta}{T} \right),
\end{equation}
where $\text{logits} \triangleq f_\theta(x, y_{<t})$ are the base logits for predicting $y_t$ given the input $x$ and prefix $y_{<t}$, $\delta \in \mathbb{R}^d$ is an additive shift vector, and $T>0$ is a temperature parameter, both learned for calibration at test time.
$W_{LM} \in \mathbb{R}^{V \times d}$ denotes the fixed language model head ($\operatorname{lm\_head}$) mapping the last hidden states of dimension $d$ to logits over a vocabulary of size $V$.
Since the model is autoregressive, this calibrated distribution adjusts the prediction of the current token and may propagate effects to future token predictions, influencing the generated sequence.

Building on this logits calibration, we design a two-phase optimization framework for test-time inference.
Specifically, we split the given inference (rollout) budget $N = N_1 + N_2$.
we first use $N_1$ to explore the output space and identify a high-scoring (high-reward) distribution, then calibrate the model’s logits to this distribution and use the remaining $N_2$ for focused exploitation (see Figure~\ref{fig:main}a).

\PYB{Up to this point, we did not formally introduce what's $T$ and $\delta$ - need similar equation as in Fig. 2} \YC{addressed.}

\PYB{Need to define $x$ and $y$} \YC{addressed.}

\noindent \textbf{Phase 1: Exploration ($N_1$ samples).}
The model generates $N_1$ candidate answers from the uncalibrated distribution $p_{\theta}(y \mid x; \delta=0, T_\mathrm{base})$, 
where $\delta=0$ and $T_\mathrm{base}$ indicate no calibration is applied in this phase.
% $x$ denotes the input question and $y$ the generated answer, $\theta$ the fixed weights of the language model, and
Each candidate is scored by a process reward model (PRM), and these scores are used to identify promising high-reward directions. 

\PYB{Need to explain the dimension of $W$} \YC{addressed. defined in first paragraph}

\noindent \textbf{Calibrate to high-scoring regions.}
From the exploration results, the top-$k$ highest-scoring completions are selected as the calibration dataset $\mathcal{D}_\mathrm{calib}(x) = \{y^{(i)}\}_{i=1}^k$.
% The calibration parameters $(\delta, T)$ are then trained directly on the cached logits of these samples, making the procedure lightweight without any additional forward passes, since learning $(\delta, T)$ is similar to shifting and rescaling the existing logits to adjust the distribution toward high-reward candidates.
The calibration parameters $(\delta, T)$ are then optimized on this problem to shift the model’s logits toward high-reward regions (see Section~\ref{method:training} for training details).

% Here, $\delta$ is a learned additive shift vector applied applied via the fixed language model head \(\operatorname{lm\_head}\), producing a token-specific bias directly to the logits: $W_\mathrm{LM} \cdot \delta$, where $W_{LM}$ was defined previously.
% To reduce dimensionality and avoid overfitting, we train $\delta$ in a lower-dimensional last hidden space $\mathbb{R}^d$ and map via $W_\mathrm{LM}$, since directly learning a full logits-space bias would be extremely high-dimensional ($V \gg d$).
% In our framework, $T$ is learned alongside $\delta$ for each input, where lower $T$ concentrates probability mass and higher $T$ flattens it.
% Together, $(\delta, T)$ provide fine-grained, interpretable control over the directly over the logits, enabling input-specific calibration without modifying the underlying model parameters.

\PYB{do we want to use another notation, like $^*$, to incidate they are optimized parameters?} \YC{added.}
\noindent \textbf{Phase 2: Calibration for Exploitation ($N_2$ samples).}
Using the learned calibration parameters $(\delta^*, T^*)$, the model generates $N_2$ candidates. These samples are focused on the high-reward regions identified during exploration, increasing the likelihood of obtaining correct or high-quality solutions.

It is noteworthy that the final answer is selected from the union of all $N_1 + N_2$ candidates, since the ground truth is unknown during inference. The exploration phase does not guarantee correctness for any individual sample, but it efficiently identifies regions in the output space that are likely to contain high-reward or plausible solutions. The exploitation phase then intensifies sampling within these regions, providing a principled balance between exploration (breadth) and exploitation (focus) under a fixed inference budget.

Let $R(x, y)$ denote the reward score assigned by the process reward model (PRM) to completion $y$ for input $x$.
The expected reward under test-time calibration can be decomposed as:

% \begin{equation}
% \mathbb{E}[R] \approx 
% \underbrace{\mathbb{E} \left[ \max_{y \in \mathcal{Y}_{\text{explore}}} R(x, y) \right]}_{\text{exploration}} 
% + 
% \underbrace{\mathbb{E} \left[ \max_{y \in \mathcal{Y}_{\text{exploit}}} \left( R(x, y) - R_{\text{explore-max}} \right)_+ \right]}_{\text{exploitation gain}}
% \end{equation}
\vspace{-0.8em}
\begin{equation}
\mathbb{E}[R_\text{final}] = \mathbb{E}\Big[ \max_{y \in \mathcal{Y}_{\text{explore}} \cup \mathcal{Y}_{\text{exploit}}} R(x, y) \Big],
\end{equation}
\vspace{-0.8em}

where $\mathcal{Y}_{\text{explore}}$ and $\mathcal{Y}_{\text{exploit}}$ are the $N_1$ exploration and $N_2$ exploitation samples, respectively.
This simple formulation highlights the exploration–exploitation tradeoff: exploration samples help cover diverse regions of the output space, while exploitation focuses on high-reward areas, jointly influencing the final maximum reward.

%\PYB{does this match the second phase where we combine both?}
%\YC{I have rewritten the formula, discarding the previous exploitation gain formulation.}

\subsection{Training $\delta$ and $T$ for Test-time Calibration}
\label{method:training}
%\YC{I have rewritten this subsection.}

% To efficiently calibrate the model toward high-reward output regions, we introduce two calibration parameters at test time: a additive shift vector $\delta \in \mathbb{R}^D$ and a temperature scaling factor $T > 0$. 
% The shift vector $\delta$ allows token-specific adjustments, while $T$ modulates the overall distribution sharpness. 
% Together, they enable fine-grained alignment toward promising outputs without modifying the base model’s parameters.
%%%

% 1. token-level calibration
\PYB{These dimensions and notations should be defined earlier; also need to explain the meaning of $C$ and $D$} \YC{I change to $d$ for hidden space and $V$ for output logits vocabulary space. alreadly mentioned in previous subsec. do we need to mention it here again?}

To guide the model toward high-reward outputs, we introduce two input-specific test-time calibration parameters: an additive shift vector $\delta \in \mathbb{R}^d$ and a temperature scaling factor $T > 0$.
For each input, the learned shift vector $\delta$ is projected through the fixed language model head $W_{LM}$ to produce a token-specific bias in logit space.
To reduce dimensionality and avoid overfitting, $\delta$ is trained in a lower-dimensional last hidden space $\mathbb{R}^d$ and mapped via $W_\mathrm{LM}$, since directly learning a full logits-space bias would be extremely high-dimensional ($V \gg d$).
The temperature $T$ is also learned for each input, providing control over the sharpness of the distribution, where lower $T$ concentrates probability mass and higher $T$ flattens it.
Together, $(\delta, T)$ provide fine-grained alignment toward high-reward completions without modifying the base model parameters.

% Here, $\delta$ is a learned additive shift vector applied applied via the fixed language model head \(\operatorname{lm\_head}\), producing a token-specific bias directly to the logits: $W_\mathrm{LM} \cdot \delta$, where $W_{LM}$ was defined previously.
% To reduce dimensionality and avoid overfitting, we train $\delta$ in a lower-dimensional last hidden space $\mathbb{R}^d$ and map via $W_\mathrm{LM}$, since directly learning a full logits-space bias would be extremely high-dimensional ($V \gg d$).
% In our framework, $T$ is learned alongside $\delta$ for each input, where lower $T$ concentrates probability mass and higher $T$ flattens it.
% Together, $(\delta, T)$ provide fine-grained, interpretable control over the directly over the logits, enabling input-specific calibration without modifying the underlying model parameters.

% At each token generation step, these parameters adjust the model’s output distribution:

% \vspace{-0.8em}
% \begin{equation}
% p_\theta(y_t \mid y_{<t}, x; \delta, T) = \operatorname{softmax} \Bigg( \frac{f(x, y_{<t}) + W_\mathrm{LM} \cdot \delta}{T} \Bigg),
% \end{equation}
% \vspace{-0.8em}

% where $f(x, y_{<t}) \in \mathbb{R}^V$ are the base logits for prefix $y_{<t}$ given input problem $x$.
% %, and $\delta \in \mathbb{R}^d$ is the learned additive calibration vector.
% This token-wise formulation is applied at every token prediction, ensuring that each token is shifted toward high-reward distributions identified during the exploration phase.

% 2. Training on Cached Logits

To efficiently learn the calibration parameters $(\delta, T)$, we leverage the top-$k$ high-reward candidates from the exploration phase, scored by the PRM, as a calibration set $\mathcal{D}_\mathrm{calib}(x) = \{y^{(i)}\}_{i=1}^k$.
Instead of repeatedly performing forward passes through the full model during calibration, we cache the base logits $f(x, y_{<t})$ for each prefix $y_{<t}$ of the high-reward candidates.
The calibration parameters $(\delta, T)$ are then optimized directly on these cached logits, making training lightweight and efficient.

\vspace{-0.8em}
\begin{equation}
(\delta^*, T^*) = \arg\min_{\delta, T>0} \ 
\mathbb{E}_{y \sim \mathcal{D}_\mathrm{calib}(x)} \left[ - \log p_\theta(y \mid x; \delta, T) \right] 
+ \lambda_\delta \|\delta\|_2^2,
\end{equation}
\vspace{-0.8em}

where $p_\theta(y \mid x) = \prod_{t=1}^{T} p_\theta(y_t \mid y_{<t}, x)$ factorizes the sequence probability over tokens, and $p_\theta(y \mid x; \delta, T)$ applies the additive shift $\delta$ and temperature $T$ at each token. 
The regularization coefficient $\lambda_\delta$ mitigates overfitting to the limited calibration set.
This calibration captures token-level effects of $(\delta, T)$, enabling input-specific adjustments under a fixed inference budget.
Further insights into the roles of $T$ and $\delta$ in calibration are provided in Sec.~\ref{discussion-insight}.
\PYB{Need to add that we provide insights into the role of $T$ and $\delta$ in calibration in Sec. XXX} \YC{added.}

\section{Theoretical Analysis}
% [lemma 1 proves that there exits a joint solution (non zero $\delta$ and $T \neq \text{start\_point}$), lemma 2 show first order approximatio solution, lemma 3 proves that there exits a unique Temp given a fixed delta shift, and theorem proves that calibration ($\delta$ and $T$) improves finite sampling's expectation reward]

%\PYB{label the theory/lemma/corr by numbers, remove section number} \YC{addressed}

Building on the previous section, we now provide theoretical foundations for test-time calibration.
We focus on answering the following question:
\textbf{RQ2. \textit{Can we provide provable guarantees that test-time calibration improves the expected reward under finite sampling?}}
To answer this, we first establish the existence of a joint calibration solution and then show that it provably improves the expected reward under Best-of-$N$ sampling.

\subsection{Existence of Joint Calibration Solutions}
% We begin by proving the existence of a joint calibration solution $(\delta, T)$ that strictly increases the probability of generating a specific high-quality output.
% The proof proceeds by construction, showing that a non-trivial shift $\delta$ and temperature $T$ can always be defined to beneficially alter the output distribution.

We begin by proving the existence of a joint calibration solution $(\delta^*, T^*)$ that strictly increases the probability of generating a high-quality output, proceeding by construction to show that a non-trivial $\delta$ and $T$ can always beneficially alter the output distribution.
\PYB{should we use $^*$ here?} \YC{added.}

\begin{lemma}[Existence of an Improving Joint Solution $(\delta, T)$]
\label{lemma-1}

Let the joint loss function be $\mathcal{L}(\delta, T) = \mathbb{E}_{y \sim \mathcal{D}_\text{calib}(x)} \left[ -\log p_\theta(y \mid x; \delta, T) \right]$.
Let $\bar{p}_{\theta}$ be the model's average predictive distribution and $\bar{p}_{\text{target}}$ be the empirical average one-hot distribution, both averaged over all generation steps in the calibration set $\mathcal{D}_\text{calib}(x)$.
Suppose the base model is not perfectly calibrated in the sense that at least one of the following conditions holds: (1) $\bar{p}_{\theta} \neq \bar{p}_{\text{target}}$, or (2) the average logit of ground-truth tokens does not equal to the average expected logit.
Then there exists a joint solution $(\delta, T) \in \mathbb{R}^D \times (0, \infty)$, where $(\delta, T) \neq (\mathbf{0}, 1)$, such that the loss is strictly reduced: $\mathcal{L}(\delta, T) < \mathcal{L}(\mathbf{0}, 1)$

\end{lemma}
\textit{Proof.} The proof is given in Appendix \ref{appendix:lemma-1}.

\subsection{Expected Reward Improvement from Calibration}

Building on the existence guarantee, we next prove that applying such a joint calibration improves the expected reward under finite Best-of-$N$ sampling.
This theorem formally establishes the benefit of our calibration method.
The proof demonstrates that the calibrated distribution achieves first-order stochastic dominance over the baseline.
Intuitively, this means the calibrated process is more likely to generate high-reward outputs, which in turn guarantees a higher expected maximum reward.

\YC{n is after calibration. n scaling insights}
\begin{theorem}[Joint Calibration $(\delta, T)$ Improves Expected Reward from Best-of-$N$ Sampling]
\label{theorem-1}

Let $p_{\theta}(y \mid x; \delta, T)$ be the model's probability distribution over outputs $y \in \mathcal{Y}$, parameterized by a calibration vector $\delta$ and a temperature $T$. Let the base model be configured with parameters $(\mathbf{0}, T_{base})$ for some $T_{base} > 0$.
Let $R(x,y)$ be a reward function, and assume there exists a unique output $y^* \in \mathcal{Y}$ with a strictly maximum reward, i.e., $r^* = R(x,y^*) > \max_{y \neq y^*} R(x,y) = r_{\text{other\_max}}$.
We consider cases where joint calibration with parameters $(\delta^*, T^*)$ improves upon the base model by increasing the probability of the unique optimal output, i.e.,
$$p_{\theta}(y^* \mid x; \delta^*, T^*) > p_{\theta}(y^* \mid x; \mathbf{0}, T_{base}).$$

Then, for any $n \geq 1$ within the remaining inference budget after calibration, the lower bound on the expected best-of-$N$ reward under the jointly calibrated model is strictly greater than that of the base model.
Specifically, let $R_{LB}(p) = r^* - (1-p)^n (r^* - r_{\text{other\_max}})$ be a valid lower bound for the expected best-of-$N$ reward, where $p$ is the probability of sampling $y^*$. The improvement in this lower bound, $\Delta_{R_{LB}}(x,n) = R_{LB}(p_{\theta}(y^* \mid x; \delta^*, T^*)) - R_{LB}(p_{\theta}(y^* \mid x; \mathbf{0}, T_{base}))$, is strictly positive.

\end{theorem}
\textit{Proof.} The proof is given in Appendix \ref{appendix:theorem-1}.

\PYB{is $n$ and $N$ the same? if so, should we just use $n$?}
\YC{TBD: $N, N_1, N_2,n$}
As a direct consequence of this theorem, we present a corollary that provides theoretical justification for our two-phase sampling strategy.
During test-time inference, one might be tempted to discard the initial exploration samples and rely only on the ``exploited'' ones. 
We show that this strategy is suboptimal, and to maximize the expected reward the final answer should be selected from the combined set of all $N_1$ exploration and $N_2$ exploitation candidates.
This result highlights that the exploration phase is essential, contributing irreplaceable value by ensuring the final candidate pool is both broad and targeted.

\begin{corollary}[Sub-optimality of Exploitation Alone]
\label{corollary-1}

The final candidate is selected by maximizing $R(x, y)$ over a set of candidates $\mathcal{Y}$. Since $\mathcal{Y}_{\text{exploit}}$ is a subset of the union $\mathcal{Y} = \mathcal{Y}_{\text{explore}} \cup \mathcal{Y}_{\text{exploit}}$, the strategy of only selecting from $\mathcal{Y}_{\text{exploit}}$ is sub-optimal compared to selecting from the union.
This is because the maximum reward achievable from the union is greater than or equal to the maximum reward achievable from the exploitation set alone.
$$
\max_{y \in \mathcal{Y}_{\text{explore}} \cup \mathcal{Y}_{\text{exploit}}} R(x, y) \geq \max_{y \in \mathcal{Y}_{\text{exploit}}} R(x, y)
$$

\end{corollary}
\textit{Proof.} The proof is given in Appendix \ref{appendix:corollary-1}.

In summary, these results affirmatively answer RQ2, showing that effective test-time calibration is achievable and beneficial.

\PYB{I think what you proved is exploitation along is no better than combined; but we didn't prove combine is optimal} \YC{exactly. i use "achievable and beneficial"}
\YC{union is not optimal, sub-optimality of exploitation only; suboptimal to union}
\YC{addressed}

\section{Results}
\subsection{Experimental Setup}

\vspace{-0.4em}
\paragraph{Models.}
% Our study considers Llama-3.2-1B-Instruct and Llama-3.2-3B-Instruct \citep{meta2024llama}, Qwen2.5-1.5B-Instruct \citep{qwen2.5} and Qwen2.5-Math-1.5B-Instruct \citep{yang2024qwen25mathtechnicalreportmathematical}, all with bf16 precision.
% These models include both general-purpose instruction-tuned LLMs (Llama and Qwen2.5) and math-specialized instruction-tuned LLMs (Qwen2.5-Math), providing a diverse set of models for evaluating calibration across different base capabilities.
% We deliberately focus on 1–3B models to balance computational feasibility with meaningful performance, 
% and all experiments start from the publicly available Hugging Face checkpoints without additional fine-tuning.

We evaluate Llama-3.2-1B/3B-Instruct \citep{meta2024llama} and Qwen2.5-1.5B-Instruct / Qwen2.5-Math-1.5B/7B-Instruct \citep{qwen2.5, yang2024qwen25mathtechnicalreportmathematical}, all in bf16.
These include general-purpose (Llama, Qwen2.5) and math-specialized (Qwen2.5-Math) models, providing diverse capabilities for calibration evaluation.
% We focus on 1–3B models to balance computation and performance, using Hugging Face checkpoints without further fine-tuning.
\vspace{-0.8em}
\paragraph{Process Reward Model (PRM).} 
% All experiments use the Qwen2.5-Math-PRM-7B \citep{zhang2025lessons}, a widely adopted PRM for mathematical reasoning. 
% It assigns step-level reward scores (0–1) to intermediate reasoning steps, enabling fine-grained evaluation beyond final answers.
% According to \citet{zhang2025lessons}, this PRM outperforms prior approaches across several benchmarks, making it the current state-of-the-art reward model for mathematical reasoning.
% Following findings in \cite{zhang2025lessons}, we adopt the reward score of the final step (\emph{last score}) as the overall score in all evaluations, since it has been shown to outperform product and minimum strategies for PRMs trained via Monte Carlo estimation.

All experiments use Qwen2.5-Math-PRM-7B \citep{zhang2025lessons}, a state-of-the-art reward model for mathematical reasoning.
It assigns step-level scores (0–1) to intermediate reasoning steps, enabling fine-grained evaluation beyond final answers.
Following \citet{zhang2025lessons}, we adopt the reward of the final step (\emph{last score}) as the overall score, which outperforms product and minimum strategies for PRMs trained via Monte Carlo estimation.

\vspace{-0.8em}
\paragraph{Baseline Setup.} 
We set $T = 0.8$ for the baseline best-of-$N$ method, as this value achieves the overall best results in a grid search over $[0.1, 1.6]$ and is consistent with previous studies \citep{snell2024scaling}. For a comprehensive analysis of temperature effects, see Appendix~\ref{appendix:llama-temp}.

\vspace{-0.8em}
\paragraph{Calibration Training.} 
% The calibration parameters $(\delta, T)$ are optimized using the AdamW optimizer in a full-batch setting for 100 epochs on each problem. 
% The learning rate is set to $0.001$ with a constant schedule. 
% A weight decay of $10^{-2}$ is applied only to $\delta$, while $T$ has no regularization. 
% The initialization is $\delta = 0$ and $T = 0.8$. 
% The loss function is the negative log-likelihood over the top-$K$ candidates. 
% For each evaluation budget $N$, we split it evenly into $N_1 = N_2 = N/2$. 
% In the first stage, we generate $N_1$ completions with $T=0.8$, and select the top-$k$ highest-scoring completions as the calibration dataset, where $k = N_1 / 4$. 
% After calibration, we generate the remaining $N_2$ completions with the calibration parameters $(\delta, T)$. 
% Notably, this procedure is lightweight because the calibration parameters $(\delta, T)$ are trained directly on cached logits from all steps, so no additional model inference is required.

Calibration parameters $(\delta, T)$ are optimized on cached logits using the top-$k$ high-scoring completions from an initial $N_1=N/2$ runs ($T=0.8$) as the calibration dataset.
The remaining $N_2=N/2$ completions are generated using the learned parameters. 
This two-stage procedure is lightweight, requires no additional inference, and learns $\delta$ and $T$ at test time for each input.
More details are provided in Appendix~\ref{appendix:calibration-training}.

\vspace{-0.8em}
\paragraph{Dataset.}
% We conduct experiments on the MATH benchmark \citep{hendrycks2021measuring}, which consists of high-school level competition math problems covering diverse topics and difficulty levels.
% For evaluation, we adopt the MATH-500 test split of 500 problems following the setup in \cite{lightman2023let}, which is widely used to assess mathematical reasoning capabilities of LLMs.
% While AIME \citep{huggingface_h4_aime_2024} is sometimes used as a benchmark, its small size (30 problems/year) and high difficulty for small LLMs mean that test-time compute yields only marginal gains. We therefore focus on MATH-500, which provides a larger and more balanced testbed for analyzing test-time calibration.
We use the MATH benchmark \citep{hendrycks2021measuring}, covering high-school level competition problems of varying topics and difficulty.
Experiments are conducted on the MATH-500 test split \citep{lightman2023let}, widely adopted for evaluating LLM mathematical reasoning.
% While AIME \citep{huggingface_h4_aime_2024} is occasionally used, its small size (30 problems/year) and high difficulty for smaller models limit test-time calibration analysis.
% MATH-500 provides a larger, balanced testbed with difficulty levels, enabling analysis of calibration across problem hardness.
Additionally, we include AIME-2024 \citep{huggingface_h4_aime_2024}, a smaller and more challenging dataset (30 problems/year), evaluated using the math-specialized Qwen2.5-Math models (1.5B and 7B).

\vspace{-0.8em}
\paragraph{Evaluation Metric.} 
% We report accuracy, defined as the proportion of generated completions whose final answers exactly match the ground-truth solutions. 
% For the \emph{vanilla} method, the accuracy is computed after selecting the single highest-scoring completion among the $N$ candidates according to scores assigned by a process reward model (PRM). 
% For the \emph{weighted} variant, we aggregate PRM scores for identical answers across candidates, summing them to obtain a weighted score for each unique answer, and then select the answer with the highest aggregated score.
Accuracy is the proportion of completions whose final answers exactly match the ground truth.
For \emph{vanilla}, the highest-scoring completion among $N$ candidates is selected.
For \emph{weighted}, PRM scores for identical answers are summed and the answer with the highest aggregated score is chosen.
All comparisons are made with the same rollout (inference) budget $N$.

\subsection{CarBoN: Calibrated Best-of-$N$ Improves Accuracy and Efficiency}

We evaluate CarBoN, which applies test-time calibration to the Best-of-$N$ strategy, on different LLMs using the MATH-500 benchmark.
We report Weighted results in Table~\ref{table-math500-weighted} (full tables including Vanilla are in Appendix~\ref{appendix:full-main-tables}), showing similar trends with greater stability across models and $N$.

For large rollout $N$ (64, 128, 256), uncalibrated Best-of-$N$ results plateau, yielding minimal gains. Llama-3.2-3B-Instruct improves only 0.6\% from $N=64$ to $256$, and Qwen2.5-1.5B-Instruct gains between 0.2\% and 0.6\%, peaking at $N=128$.
In contrast, CarBoN continues to improve performance beyond this limit.
For example, all models achieve higher accuracy at $N=64$ with CarBoN than the uncalibrated baseline at $N=256$, reducing the required rollout budgets by up to 4$\times$.
Notably, Qwen2.5-Math-1.5B-Instruct with CarBoN at $N=64$ reaches 77.2\% accuracy, surpassing GPT-4o @1 (77.0\%; see Appendix~\ref{appendix:bigger_llm} for more details and results on larger models), while the uncalibrated Best-of-$N$ at $N=256$ reaches only 76.8\%.

At $N=256$, CarBoN improves Weighted decoding by over 1\% across all models. Vanilla decoding also shows notable gains, up to 3.6\% for Qwen2.5-1.5B-Instruct. These results show that CarBoN not only increases accuracy but also reduces sampling costs.

We also experiment with larger math-specialized models (Qwen2.5-Math-7B-Instruct) and a more challenging benchmark, AIME-2024 \citep{huggingface_h4_aime_2024}, which contains 30 high-difficulty problems. 
Table~\ref{table-aime-weighted} reports the number of correct answers for both Qwen2.5-Math-1.5B/7B-Instruct across different rollout budgets $N$. 
Even on this small and difficult dataset, CarBoN improves over the uncalibrated Best-of-$N$, demonstrating that test-time calibration boosts performance for larger models and harder problems while requiring fewer rollouts.

%We evaluate both Qwen2.5-Math-1.5B-Instruct and the larger Qwen2.5-Math-7B-Instruct on theAIME-2024\citep{huggingface_h4_aime_2024}.
%Table~\ref{table-aime-weighted} shows that CarBoN improves performance over uncalibrated Best-of-$N$ for both model sizes, demonstrating its effectiveness for larger models and harder problems while requiring fewer rollouts.

Overall, these results show that CarBoN, which is a concrete instance of the test-time calibration framework, consistently improves reasoning performance and enhances the quality of selected outputs without modifying the underlying decoding strategy.

\begin{table}[t]
\centering
\caption{
\textbf{Accuracy (\%) of four models on MATH-500, comparing Weighted Best-of-$N$ methods before and after calibration.}
CarBoN enables further improvements beyond the plateau of standard Best-of-$N$, with calibrated accuracy at $N=64$ exceeding the uncalibrated results at $N=256$, corresponding to up to $4\times$ less rollout budgets.
Bold indicates better accuracy for each $N$. 
}
\vspace{-0.4em}
\resizebox{0.8\textwidth}{!}{
\begin{tabular}{ccrrrrrr}
\toprule
\multirow{2}{*}{Model} & \multirow{2}{*}{Method} & \multicolumn{6}{c}{N} \\  [-1.0mm]
\cmidrule(lr){3-8} 
 & & \multicolumn{1}{c}{8} & \multicolumn{1}{c}{16} & \multicolumn{1}{c}{32} & \multicolumn{1}{c}{64} & \multicolumn{1}{c}{128} & \multicolumn{1}{c}{256} \\  [-0.8mm]
\midrule
\multirow{2}{*}{Llama-3.2-1B-Instruct}      
& Best-of-$N$ & 42.0 & 44.6 & 47.8 & 48.6 & 50.6 & 50.8 \\
& CarBoN    & \textbf{43.0} & \textbf{45.6} & \textbf{48.4} & \textbf{51.0} & \textbf{51.8} & \textbf{51.8} \\
\midrule
\multirow{2}{*}{Llama-3.2-3B-Instruct}      
& Best-of-$N$ & 56.8 & 58.2 & 59.6 & 61.6 & 61.8 & 62.2 \\
& CarBoN    & \textbf{57.6} & \textbf{59.0} & \textbf{60.8} & \textbf{62.2} & \textbf{63.2} & \textbf{63.4} \\
\midrule
\multirow{2}{*}{Qwen2.5-1.5B-Instruct}      
& Best-of-$N$ & \textbf{56.4} & 57.6 & 61.4 & 62.0 & 62.6 & 62.2 \\
& CarBoN    & 55.0 & \textbf{60.0} & \textbf{61.8} & \textbf{62.4} & \textbf{64.0} & \textbf{64.4} \\
\midrule
\multirow{2}{*}{Qwen2.5-Math-1.5B-Instruct} 
& Best-of-$N$ & 73.6 & 75.4 & \textbf{76.4} & 75.6 & 76.4 & 76.8 \\
& CarBoN    & \textbf{74.2} & \textbf{76.0} & \textbf{76.4} & \textbf{77.2} & \textbf{77.2} & \textbf{77.8} \\
\bottomrule
\end{tabular}
}
\label{table-math500-weighted}
\end{table}
\vspace{-1mm}
\begin{table}[ht]
\centering
\caption{
\textbf{Correct answers (out of 30) on the AIME-2024 benchmark for two math-specialized models, comparing Best-of-$N$ and CarBoN across different rollout budgets.} 
CarBoN enables further improvements beyond the plateau of standard Best-of-$N$.
Bold numbers indicate the higher number of correct answers for each $N$.
}
\label{table-aime-weighted}
\vspace{-0.4em}
\resizebox{0.8\textwidth}{!}{
\begin{tabular}{c c r r r r r}
\toprule
\multirow{2}{*}{Model} & \multirow{2}{*}{Method} & \multicolumn{5}{c}{N} \\  [-1.0mm]
\cmidrule(lr){3-7} 
 & & \multicolumn{1}{c}{16} & \multicolumn{1}{c}{32} & \multicolumn{1}{c}{64} & \multicolumn{1}{c}{128} & \multicolumn{1}{c}{256} \\  [-0.8mm]
\midrule
\multirow{2}{*}{Qwen2.5-Math-1.5B-Instruct} 
& Best-of-$N$ & 4/30 & 5/30 & 6/30 & 6/30 & 6/30 \\
& CarBoN & 4/30 & 5/30 & 6/30 & \textbf{7/30} & \textbf{7/30} \\
\midrule
\multirow{2}{*}{Qwen2.5-Math-7B-Instruct} 
& Best-of-$N$ & 5/30 & 5/30 & 6/30 & 6/30 & 6/30 \\
& CarBoN & 5/30 & \textbf{6/30} & 6/30 & 6/30 & \textbf{7/30} \\
\bottomrule
\end{tabular}
}
\end{table}
\vspace{-1mm}

% For example, all models achieve higher accuracy at $N=64$ with CarBoN than the uncalibrated baseline at $N=256$, reducing required test-time compute by up to 4$\times$.
% Notably, as shown in the rightmost line graph of Figure~\ref{fig:main} (b), Qwen2.5-Math-1.5B-Instruct with CarBoN at $N=64$ reaches 77.2\% accuracy, surpassing GPT-4o @1 (76.9\%), while the uncalibrated Best-of-$N$ at $N=256$ reaches only 76.8\%.
% At $N=256$, CarBoN improves Weighted decoding by over 1\% across all models, while Vanilla decoding benefits even more, with the largest gain of 3.6\% for Qwen2.5-1.5B-Instruct.
% These results indicate that CarBoN not only increases accuracy but also significantly enhances efficiency, particularly when sampling costs are high.

\subsection{Effect of the Calibration Parameters $(\delta, T)$}

% We perform an ablation study on a general-purpose model (Llama-3.2-1B-Instruct) and a math-specialist model (Qwen2.5-Math-1.5B-Instruct) to analyze the individual contributions of the additive shift $\delta$ and temperature scaling $T$.
% For each experiment, we retrain calibration while enabling only one parameter at a time, ensuring that the observed effects reflect the true impact of $\delta$ or $T$.

% Table~\ref{table-ablation} shows different variants across models. For Llama-3.2-1B-Instruct, either $\delta$ or $T$ alone sometimes achieves the highest score for certain $N$ or naive-selection cases, such as $\delta$ at $N=32$ (48.8\%) and $T$ at $N=16$ (47.6\%).
% The naive selection occasionally favors a single calibration, which explains why CarBoN does not always yield the maximum score in these settings. However, when using weighted selection, CarBoN consistently outperforms all alternatives, reaching up to 51.8\% at $N=128$ and $256$.

% For Qwen2.5-Math-1.5B-Instruct, CarBoN dominates across nearly all $N$ and both selection strategies, achieving the highest scores in almost every case (up to 77.8\% at $N=256$ with weighted selection). This indicates that $\delta$ and $T$ provide complementary improvements in this domain-specialized setting. Overall, the combination of both parameters is the most reliable way to boost performance, particularly with weighted selection.

We perform an ablation study on a general-purpose model (Llama-3.2-1B-Instruct) and a math-specialist model (Qwen2.5-Math-1.5B-Instruct) to isolate the contributions of the additive shift $\delta$ and temperature $T$ (full tables including Vanilla decoding are in Appendix~\ref{appendix:full-main-tables}).
For each experiment, we retrain calibration with only one parameter enabled, isolating the effect of $\delta$ or $T$.

Table~\ref{table-ablation-weighted} shows that adding $\delta$ alone already improves over the baseline once $N$ is sufficiently large, while the combination of $\delta$ and $T$ (CarBoN) yields the strongest gains.
For instance, CarBoN reaches 51.8\% for Llama-3.2-1B-Instruct and 77.8\% for Qwen2.5-Math-1.5B-Instruct.
These results indicate that using $\delta$ and $T$ together provides the most reliable gains.
In the next section, we analyze how $\delta$ and $T$ contribute to answer quality.

\begin{table}[b]
\centering
\caption{
\textbf{Ablation study on calibration parameters $(\delta, T)$ and their combination (CarBoN) for Best-of-$N$ search on MATH-500.}
We compare applying a shift ($\delta$), a temperature scaling ($T$), and their joint calibration (CarBoN) under Weighted selection.
All values report accuracy (\%).
Results show that CarBoN consistently improves accuracy across different $N$, highlighting the complementary benefits of $\delta$ and $T$.
Bold numbers indicate the better accuracy for each $N$.
}
\vspace{-0.5em}
\resizebox{0.8\textwidth}{!}{
\begin{tabular}{cclrrrrrr}
\toprule
\multirow{2}{*}{Model} & \multirow{2}{*}{Method} & \multicolumn{6}{c}{N} \\  [-1.0mm]
\cmidrule(lr){3-8} 
 & & \multicolumn{1}{c}{8} & \multicolumn{1}{c}{16} & \multicolumn{1}{c}{32} & \multicolumn{1}{c}{64} & \multicolumn{1}{c}{128} & \multicolumn{1}{c}{256} \\  [-0.8mm]
\midrule
\multirow{4}{*}{Llama-3.2-1B-Instruct}      
& Best-of-$N$        & 42.0 & 44.6 & 47.8 & 48.6 & 50.6 & 50.8 \\
& Best-of-$N$ w/ $\delta$ & 41.8 & 43.8 & 48.2 & 49.0 & 51.0 & 51.2 \\
& Best-of-$N$ w/ $T$      & 42.0 & 45.0 & 47.0 & 49.6 & 49.8 & 50.6 \\
& CarBoN                & \textbf{43.0} & \textbf{45.6} & \textbf{48.4} & \textbf{51.0} & \textbf{51.8} & \textbf{51.8} \\
\midrule
\multirow{4}{*}{Qwen2.5-Math-1.5B-Instruct} 
& Best-of-$N$        & 73.6 & 75.4 & \textbf{76.4} & 75.6 & 76.4 & 76.8 \\
& Best-of-$N$ w/ $\delta$ & \textbf{74.2} & 74.8 & 75.6 & 77.0 & 76.6 & 77.0 \\
& Best-of-$N$ w/ $T$      & 73.2 & 75.2 & 76.0 & 76.4 & 76.0 & 76.6 \\
& CarBoN                & \textbf{74.2} & \textbf{76.0} & \textbf{76.4} & \textbf{77.2} & \textbf{77.2} & \textbf{77.8} \\
\bottomrule
\end{tabular}
}
\label{table-ablation-weighted}
\end{table}
\vspace{-1mm}

\section{Discussion of Calibration and Generalization}

Beyond the main results and ablation studies, we further analyze the distinct roles of temperature $T$ and $\delta$, and examine the generalization of test-time calibration beyond Best-of-$N$ sampling.

\subsection{How Token-level Calibration Improves Answer Quality}
\label{discussion-insight}
\begin{figure*}[t]
    \centering
    \includegraphics[width=\linewidth]{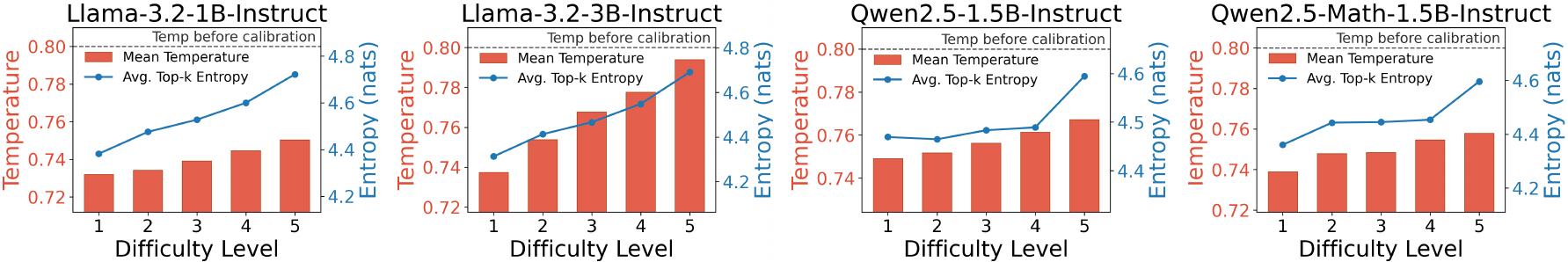}
    \vspace{-0.8em}
    \caption{\textbf{Correlation between problem difficulty, calibrated temperature, and top-$k$ completion entropy on MATH-500.} Bars (left y-axis) show the average learned temperature across five difficulty levels, while the line plot (right y-axis) shows the normalized entropy of the top-$k$ completions used for calibration. Both temperature and entropy strongly increase with problem difficulty, indicating that harder problems require higher temperatures to capture the more diverse top-$k$ token distributions. See Appendix~\ref{appendix:cor_level_temp} for full Spearman correlation statistics.}
    \vspace{-0.4em}
    \label{fig:level_vs_temp}
\end{figure*}

\vspace{-0.3em}
\paragraph{Temperature Adaptation.}
We find that calibration temperature strongly correlates with problem difficulty.
In Figure \ref{fig:level_vs_temp} (with $N_1=128$ for exploration, $k=32$ for calibration), harder questions have higher temperatures, since greater diversity improves the chance of reaching correct answers.
To explain this, we analyze the entropy of the top-$k$ high-scoring completions used for calibration.
The blue curve shows entropy rising with difficulty, meaning the model produces more diverse outputs when less confident.
Although calibration only uses high-scoring responses, this diversity remains, enabling the model to learn an appropriate temperature for different difficulty levels.

Beyond difficulty, we also find that the calibration temperature increases with $N$, as higher temperatures promote diversity and better utilize the inference budget (see Appendix~\ref{appendix:temp_vs_n}).
Larger $N$ enables more exploration, while a higher temperature prevents near-identical samples, ensuring that the additional budget contributes meaningfully.
This highlights that temperature should adapt to problem difficulty and inference budget, and our calibration achieves this adaptation.

\vspace{-0.3em}
\paragraph{Delta Adjustment.}

\begin{wraptable}[9]{r}{0.45\textwidth}
\centering
% \begin{table}[t]
\centering
\vspace{-1em}
\captionof{table}{Token-level overlap with/without $\delta$ calibration against top-$k$ high-scoring answers on Llama-3.2-1B-Instruct ($N_1=128, k=32$).}
\label{table-overlap-delta}
\vspace{-0.5em}
\resizebox{0.45\textwidth}{!}{
\begin{tabular}{lcc}
\toprule
Metric & Calibration w/ $\delta$ & No Calibration \\ [-0.5mm]
\midrule
Jaccard Overlap $\uparrow$   & \textbf{0.5184} & 0.4838 \\
Dice Overlap $\uparrow$      & \textbf{0.6805} & 0.6497 \\
Token Recall $\uparrow$      & 0.8804 & \textbf{0.8919} \\
Token Precision $\uparrow$   & \textbf{0.5590} & 0.5147 \\
\bottomrule
\end{tabular}
}
% \end{table}
\end{wraptable}
We report four overlap metrics with respect to top-$k$ high-scoring answers: Jaccard and Dice (set-level similarity), and Recall and Precision (token-level coverage and specificity), with full definitions in the Appendix~\ref{appendix:delta_token_metric}.
After applying $\delta$ calibration, the generated tokens show higher set-level similarity, as measured by Jaccard and Dice, and increased token-level Precision, with the results for each metric comparing the calibrated and uncalibrated settings presented in Table~\ref{table-overlap-delta}, indicating that the model’s generations are more aligned with the patterns of high-quality responses.
These results are consistent with our ablation study in Table~\ref{table-ablation-weighted}, where incorporating $\delta$ improves task-level accuracy, particularly when aggregating a larger candidate pool, suggesting that $\delta$ effectively guides the model toward high-scoring behavior.
\PYB{k or $k$? need to be consistent} \YC{addressed. it's $k$.}

\subsection{Generalizing Test-time Calibration Beyond Best-of-$N$}
While test-time calibration improves reasoning ability in Best-of-$N$ and offers greater efficiency, other step-level strategies score  each reasoning step before proceeding to the next.
While such methods (e.g., beam search, DVTS, and particle filtering) can achieve higher performance when more fine-grained guidance is available during generation, they need  heavier computation due to repeated verifier calls.
To illustrate that test-time calibration can generalize beyond Best-of-$N$, we focus on beam search as a representative step-level sampling strategy.

As shown in Table~\ref{table-beam-search}, calibrated beam search provides improvements over the standard baseline in most settings across both models.
Notably, with $N=32$, the calibrated beam search reaches accuracy close to or even matching that of beam search with $N=64$, indicating that calibration improves sample efficiency by reducing the number of candidates required to achieve a given performance level.
This demonstrates that test-time calibration is not limited to Best-of-$N$ but can also enhance fine-grained step-level decoding, suggesting a promising direction for integrating calibration with step-level sampling methods.

\begin{table}[ht]
\centering
\caption{
\textbf{Accuracy (\%) of standard and calibrated beam search on the MATH-500 benchmark.}  
Calibrated beam search generally improves test-time reasoning performance, especially for larger $N$.  
}
\vspace{-0.4em}
\resizebox{0.8\textwidth}{!}{

\begin{tabular}{cclrrrr}
\toprule
\multirow{2}{*}{Model}                      & \multicolumn{2}{c}{\multirow{2}{*}{Method}} & \multicolumn{4}{c}{N}                                                                            \\ [-1.0mm]
\cmidrule(lr){4-7}
                                            & \multicolumn{2}{c}{}                        & \multicolumn{1}{c}{8} & \multicolumn{1}{c}{16} & \multicolumn{1}{c}{32} & \multicolumn{1}{c}{64} \\ [-0.8mm]
\midrule
\multirow{2}{*}{Llama-3.2-1B-Instruct}      & \multicolumn{2}{c}{Beam Search}             & 56.0                & 58.4                 & 60.4                 & 62.2                 \\ \cmidrule(lr){2-7}
                                            & \multicolumn{2}{c}{Calibrated Beam Search}  & \textbf{57.2}       & \textbf{60.0}        & \textbf{62.2}        & \textbf{64.2}        \\
\midrule
\multirow{2}{*}{Qwen2.5-Math-1.5B-Instruct} & \multicolumn{2}{c}{Beam Search}             & \textbf{79.0}       & 79.2                 & 80.2                 & 81.4                 \\ \cmidrule(lr){2-7}
                                            & \multicolumn{2}{c}{Calibrated Beam Search}  & 78.6                & \textbf{79.6}        & \textbf{81.2}        & \textbf{82.8}        \\
\bottomrule
\end{tabular}

}
\label{table-beam-search}
\end{table}
\section{Conclusion}

We introduced test-time calibration, a framework to adapt LLMs at inference under test-time scaling via additive logits shifts and adaptive temperature scaling, instantiated as CarBoN on Best-of-N.
Our theoretical analysis shows calibration can provably improve accuracy and the lower bound of expected reward under finite samples.
Empirically, CarBoN consistently improves performance across benchmarks, rollout budgets, and step-wise sampling, demonstrating its generalization potential.
We believe this framework will inspire and advance future designs of test-time scaling methods.

% \PYB{believe}

\newpage
\bibliography{ref}

\begin{thebibliography}{30}
\providecommand{\natexlab}[1]{#1}
\providecommand{\url}[1]{\texttt{#1}}
\expandafter\ifx\csname urlstyle\endcsname\relax
  \providecommand{\doi}[1]{doi: #1}\else
  \providecommand{\doi}{doi: \begingroup \urlstyle{rm}\Url}\fi

\bibitem[{Anthropic}(2025)]{anthropic2025tracing}
{Anthropic}.
\newblock Tracing the thoughts of a large language model.
\newblock \url{https://www.anthropic.com/research/tracing-thoughts-language-model}, 2025.
\newblock Accessed: 2025-09-10.

\bibitem[Beeching et~al.()Beeching, Tunstall, and Rush]{beeching2024scalingtesttimecompute}
Edward Beeching, Lewis Tunstall, and Sasha Rush.
\newblock Scaling test-time compute with open models.
\newblock URL \url{https://huggingface.co/spaces/HuggingFaceH4/blogpost-scaling-test-time-compute}.

\bibitem[Brown et~al.(2024)Brown, Juravsky, Ehrlich, Clark, Le, R{\'e}, and Mirhoseini]{brown2024large}
Bradley Brown, Jordan Juravsky, Ryan Ehrlich, Ronald Clark, Quoc~V Le, Christopher R{\'e}, and Azalia Mirhoseini.
\newblock Large language monkeys: Scaling inference compute with repeated sampling.
\newblock \emph{arXiv preprint arXiv:2407.21787}, 2024.

\bibitem[Guo et~al.(2017)Guo, Pleiss, Sun, and Weinberger]{guo2017calibration}
Chuan Guo, Geoff Pleiss, Yu~Sun, and Kilian~Q Weinberger.
\newblock On calibration of modern neural networks.
\newblock In \emph{International conference on machine learning}, pp.\  1321--1330. PMLR, 2017.

\bibitem[Guo et~al.(2025)Guo, Yang, Zhang, Song, Zhang, Xu, Zhu, Ma, Wang, Bi, et~al.]{guo2025deepseek}
Daya Guo, Dejian Yang, Haowei Zhang, Junxiao Song, Ruoyu Zhang, Runxin Xu, Qihao Zhu, Shirong Ma, Peiyi Wang, Xiao Bi, et~al.
\newblock Deepseek-r1: Incentivizing reasoning capability in llms via reinforcement learning.
\newblock \emph{arXiv preprint arXiv:2501.12948}, 2025.

\bibitem[Hendrycks et~al.(2021)Hendrycks, Burns, Kadavath, Arora, Basart, Tang, Song, and Steinhardt]{hendrycks2021measuring}
Dan Hendrycks, Collin Burns, Saurav Kadavath, Akul Arora, Steven Basart, Eric Tang, Dawn Song, and Jacob Steinhardt.
\newblock Measuring mathematical problem solving with the math dataset.
\newblock \emph{arXiv preprint arXiv:2103.03874}, 2021.

\bibitem[{HuggingFaceH4}(2024)]{huggingface_h4_aime_2024}
{HuggingFaceH4}.
\newblock Huggingfaceh4/aime\_2024.
\newblock \url{https://huggingface.co/datasets/HuggingFaceH4/aime_2024}, 2024.
\newblock A dataset consisting of 30 problems from AIME I and II 2024.

\bibitem[Kojima et~al.(2022)Kojima, Gu, Reid, Matsuo, and Iwasawa]{kojima2022large}
Takeshi Kojima, Shixiang~Shane Gu, Machel Reid, Yutaka Matsuo, and Yusuke Iwasawa.
\newblock Large language models are zero-shot reasoners.
\newblock \emph{Advances in neural information processing systems}, 35:\penalty0 22199--22213, 2022.

\bibitem[Kull et~al.(2019)Kull, Perello~Nieto, K{\"a}ngsepp, Silva~Filho, Song, and Flach]{kull2019beyond}
Meelis Kull, Miquel Perello~Nieto, Markus K{\"a}ngsepp, Telmo Silva~Filho, Hao Song, and Peter Flach.
\newblock Beyond temperature scaling: Obtaining well-calibrated multi-class probabilities with dirichlet calibration.
\newblock \emph{Advances in neural information processing systems}, 32, 2019.

\bibitem[Lightman et~al.(2023)Lightman, Kosaraju, Burda, Edwards, Baker, Lee, Leike, Schulman, Sutskever, and Cobbe]{lightman2023let}
Hunter Lightman, Vineet Kosaraju, Yuri Burda, Harrison Edwards, Bowen Baker, Teddy Lee, Jan Leike, John Schulman, Ilya Sutskever, and Karl Cobbe.
\newblock Let's verify step by step.
\newblock In \emph{The Twelfth International Conference on Learning Representations}, 2023.

\bibitem[Liu et~al.(2025)Liu, Gao, Zhao, Zhang, Li, Qi, Ouyang, and Zhou]{liu2025can}
Runze Liu, Junqi Gao, Jian Zhao, Kaiyan Zhang, Xiu Li, Biqing Qi, Wanli Ouyang, and Bowen Zhou.
\newblock Can 1b llm surpass 405b llm? rethinking compute-optimal test-time scaling.
\newblock \emph{arXiv preprint arXiv:2502.06703}, 2025.

\bibitem[{Meta AI}(2024)]{meta2024llama}
{Meta AI}.
\newblock Llama 3.2: Revolutionizing edge ai and vision with open, customizable models, 2024.
\newblock URL \url{https://ai.meta.com/blog/llama-3-2-connect-2024-vision-edge-mobile-devices/}.
\newblock Accessed: 2025-08-06.

\bibitem[OpenAI(2024)]{openai2024learning}
OpenAI.
\newblock Learning to reason with llms, 2024.
\newblock URL \url{https://openai.com/index/learning-to-reason-with-llms/}.
\newblock Accessed: 2025-08-06.

\bibitem[Puri et~al.(2025)Puri, Sudalairaj, Xu, Xu, and Srivastava]{puri2025probabilistic}
Isha Puri, Shivchander Sudalairaj, Guangxuan Xu, Kai Xu, and Akash Srivastava.
\newblock A probabilistic inference approach to inference-time scaling of llms using particle-based monte carlo methods.
\newblock \emph{arXiv preprint arXiv:2502.01618}, 2025.

\bibitem[Qu et~al.(2024)Qu, Zhang, Garg, and Kumar]{qu2024recursive}
Yuxiao Qu, Tianjun Zhang, Naman Garg, and Aviral Kumar.
\newblock Recursive introspection: Teaching language model agents how to self-improve.
\newblock \emph{Advances in Neural Information Processing Systems}, 37:\penalty0 55249--55285, 2024.

\bibitem[{Qwen Team}(2024)]{qwen2.5}
{Qwen Team}.
\newblock Qwen2.5: A party of foundation models, September 2024.
\newblock URL \url{https://qwenlm.github.io/blog/qwen2.5/}.

\bibitem[Setlur et~al.(2025)Setlur, Rajaraman, Levine, and Kumar]{setlur2025scaling}
Amrith Setlur, Nived Rajaraman, Sergey Levine, and Aviral Kumar.
\newblock Scaling test-time compute without verification or rl is suboptimal.
\newblock \emph{arXiv preprint arXiv:2502.12118}, 2025.

\bibitem[Shinn et~al.(2023)Shinn, Cassano, Gopinath, Narasimhan, and Yao]{shinn2023reflexion}
Noah Shinn, Federico Cassano, Ashwin Gopinath, Karthik Narasimhan, and Shunyu Yao.
\newblock Reflexion: Language agents with verbal reinforcement learning.
\newblock \emph{Advances in Neural Information Processing Systems}, 36:\penalty0 8634--8652, 2023.

\bibitem[Snell et~al.(2024)Snell, Lee, Xu, and Kumar]{snell2024scaling}
Charlie Snell, Jaehoon Lee, Kelvin Xu, and Aviral Kumar.
\newblock Scaling llm test-time compute optimally can be more effective than scaling model parameters.
\newblock \emph{arXiv preprint arXiv:2408.03314}, 2024.

\bibitem[Tang et~al.(2024)Tang, Chen, and Ho]{tang2024neural}
Yung-Chen Tang, Pin-Yu Chen, and Tsung-Yi Ho.
\newblock Neural clamping: Joint input perturbation and temperature scaling for neural network calibration.
\newblock \emph{Transactions on Machine Learning Research}, 2024.
\newblock ISSN 2835-8856.
\newblock URL \url{https://openreview.net/forum?id=qSFToMqLcq}.

\bibitem[Wang et~al.(2023)Wang, Li, Shao, Xu, Dai, Li, Chen, Wu, and Sui]{wang2023math}
Peiyi Wang, Lei Li, Zhihong Shao, RX~Xu, Damai Dai, Yifei Li, Deli Chen, Yu~Wu, and Zhifang Sui.
\newblock Math-shepherd: Verify and reinforce llms step-by-step without human annotations.
\newblock \emph{arXiv preprint arXiv:2312.08935}, 2023.

\bibitem[Wang et~al.(2022)Wang, Wei, Schuurmans, Le, Chi, Narang, Chowdhery, and Zhou]{wang2022self}
Xuezhi Wang, Jason Wei, Dale Schuurmans, Quoc Le, Ed~Chi, Sharan Narang, Aakanksha Chowdhery, and Denny Zhou.
\newblock Self-consistency improves chain of thought reasoning in language models.
\newblock \emph{arXiv preprint arXiv:2203.11171}, 2022.

\bibitem[Wei et~al.(2022)Wei, Wang, Schuurmans, Bosma, Xia, Chi, Le, Zhou, et~al.]{wei2022chain}
Jason Wei, Xuezhi Wang, Dale Schuurmans, Maarten Bosma, Fei Xia, Ed~Chi, Quoc~V Le, Denny Zhou, et~al.
\newblock Chain-of-thought prompting elicits reasoning in large language models.
\newblock \emph{Advances in neural information processing systems}, 35:\penalty0 24824--24837, 2022.

\bibitem[Yang et~al.(2024)Yang, Zhang, Hui, Gao, Yu, Li, Liu, Tu, Zhou, Lin, Lu, Xue, Lin, Liu, Ren, and Zhang]{yang2024qwen25mathtechnicalreportmathematical}
An~Yang, Beichen Zhang, Binyuan Hui, Bofei Gao, Bowen Yu, Chengpeng Li, Dayiheng Liu, Jianhong Tu, Jingren Zhou, Junyang Lin, Keming Lu, Mingfeng Xue, Runji Lin, Tianyu Liu, Xingzhang Ren, and Zhenru Zhang.
\newblock Qwen2.5-math technical report: Toward mathematical expert model via self-improvement.
\newblock \emph{arXiv preprint arXiv:2409.12122}, 2024.

\bibitem[Yu et~al.(2023)Yu, Jiang, Shi, Yu, Liu, Zhang, Kwok, Li, Weller, and Liu]{yu2023metamath}
Longhui Yu, Weisen Jiang, Han Shi, Jincheng Yu, Zhengying Liu, Yu~Zhang, James~T Kwok, Zhenguo Li, Adrian Weller, and Weiyang Liu.
\newblock Metamath: Bootstrap your own mathematical questions for large language models.
\newblock \emph{arXiv preprint arXiv:2309.12284}, 2023.

\bibitem[Yue et~al.(2023)Yue, Qu, Zhang, Fu, Huang, Sun, Su, and Chen]{yue2023mammoth}
Xiang Yue, Xingwei Qu, Ge~Zhang, Yao Fu, Wenhao Huang, Huan Sun, Yu~Su, and Wenhu Chen.
\newblock Mammoth: Building math generalist models through hybrid instruction tuning.
\newblock \emph{arXiv preprint arXiv:2309.05653}, 2023.

\bibitem[Zadrozny \& Elkan(2001)Zadrozny and Elkan]{zadrozny2001obtaining}
Bianca Zadrozny and Charles Elkan.
\newblock Obtaining calibrated probability estimates from decision trees and naive bayesian classifiers.
\newblock In \emph{ICML}, volume~1, 2001.

\bibitem[Zadrozny \& Elkan(2002)Zadrozny and Elkan]{zadrozny2002transforming}
Bianca Zadrozny and Charles Elkan.
\newblock Transforming classifier scores into accurate multiclass probability estimates.
\newblock In \emph{Proceedings of the eighth ACM SIGKDD international conference on Knowledge discovery and data mining}, pp.\  694--699, 2002.

\bibitem[Zhang et~al.(2025)Zhang, Zheng, Wu, Zhang, Lin, Yu, Liu, Zhou, and Lin]{zhang2025lessons}
Zhenru Zhang, Chujie Zheng, Yangzhen Wu, Beichen Zhang, Runji Lin, Bowen Yu, Dayiheng Liu, Jingren Zhou, and Junyang Lin.
\newblock The lessons of developing process reward models in mathematical reasoning.
\newblock \emph{arXiv preprint arXiv:2501.07301}, 2025.

\bibitem[Zhou et~al.(2022)Zhou, Sch{\"a}rli, Hou, Wei, Scales, Wang, Schuurmans, Cui, Bousquet, Le, et~al.]{zhou2022least}
Denny Zhou, Nathanael Sch{\"a}rli, Le~Hou, Jason Wei, Nathan Scales, Xuezhi Wang, Dale Schuurmans, Claire Cui, Olivier Bousquet, Quoc Le, et~al.
\newblock Least-to-most prompting enables complex reasoning in large language models.
\newblock \emph{arXiv preprint arXiv:2205.10625}, 2022.

\end{thebibliography}
\bibliographystyle{iclr2026_conference}

\clearpage
\appendix

\section*{LLM Usage Disclosure}
We used Large Language Models to assist in writing and coding for this paper.
ChatGPT and Gemini were employed to help polish language, improve clarity, and refine expression.
GitHub Copilot was used to provide autocomplete suggestions and minor code snippets.
All core ideas, designs, and conclusions were independently developed and verified by the authors.

\section*{Reproducibility Statement}
To ensure reproducibility, we provide detailed descriptions of our methodology, theoretical derivations, and experimental setup in the paper and appendix. The code used for experiments is included in the supplementary materials to support verification and replication of our results.

\section{Mathematical Proofs}
\subsection{Proof of Lemma ~\ref{lemma-1}}
\label{appendix:lemma-1}
\textbf{Lemma ~\ref{lemma-1}} (Existence of an Improving Joint Solution $(\delta, T)$)
\begin{itshape}
Let the joint loss function be $\mathcal{L}(\delta, T) = \mathbb{E}_{y \sim \mathcal{D}_\text{calib}(x)} \left[ -\log p_\theta(y \mid x; \delta, T) \right]$.
Let $\bar{p}_{\theta}$ be the model's average predictive distribution and $\bar{p}_{\text{target}}$ be the empirical average one-hot distribution, both averaged over all generation steps in the calibration set $\mathcal{D}_\text{calib}(x)$.
Suppose the base model is not perfectly calibrated in the sense that at least one of the following conditions holds: (1) $\bar{p}_{\theta} \neq \bar{p}_{\text{target}}$, or (2) the average logit of ground-truth tokens does not equal to the average expected logit.
Then there exists a joint solution $(\delta, T) \in \mathbb{R}^D \times (0, \infty)$, where $(\delta, T) \neq (\mathbf{0}, 1)$, such that the loss is strictly reduced:$$\mathcal{L}(\delta, T) < \mathcal{L}(\mathbf{0}, 1)$$
\end{itshape}

%%% Previous lemma statement
% Let the joint loss function be $\mathcal{L}(\delta, T) = \mathbb{E}_{y \sim \mathcal{D}_\text{calib}(x)} \left[ -\log p_\theta(y \mid x; \delta, T) \right]$.
% Suppose the base model is not perfectly calibrated to the empirical distribution over the calibration set $\mathcal{D}_\text{calib}(x)$, i.e., $p_\theta(\ \cdot \mid x) \neq p_{\text{target}}(\ \cdot \mid x)$ where $p_{\text{target}}(\ \cdot \mid x)$ denotes the empirical distribution over $\mathcal{D}_\text{calib}(x)$.
% Then there exists a joint solution $(\delta, T) \in \mathbb{R}^D \times (0, \infty)$, where $(\delta, T) \neq (\mathbf{0}, 1)$, such that the loss is strictly reduced:$$\mathcal{L}(\delta, T) < \mathcal{L}(\mathbf{0}, 1)$$

\begin{proof}

The proof demonstrates that the joint loss function $\mathcal{L}(\delta, T)$ is continuously differentiable and that its gradient, evaluated at the initial point $(\delta, T)=(\mathbf{0}, 1)$, is a non-zero vector. For a continuously differentiable function, a non-zero gradient at a point guarantees the existence of a strict descent direction, ensuring that a nearby point with a lower loss value exists.

\paragraph{Continuity and Differentiability.}
The loss $\mathcal{L}(\delta, T)$ is the average of the per-step negative log-likelihoods (NLL) $L_{i,j}(\delta, T)$ over the calibration set.
The per-step NLL is a function of the logits, which are an affine transformation of $\delta$, divided by temperature $T$, and then passed through a log-softmax function.
Specifically, $L_{i,j}(\delta, T) = -\log(\text{softmax}((W_{\text{lm}}(h_{i,j}+\delta))/T))_{y_{i,j}}$. Since affine transformations, division by a non-zero scalar, the exponential function, and the logarithm function are all continuously differentiable ($C^\infty$) on their domains, their composition, the log-softmax, is also continuously differentiable. As $\mathcal{L}$ is a finite sum of such functions, it is also continuously differentiable on its domain $\mathbb{R}^D \times (0, \infty)$.

\paragraph{The Joint Loss Function.}
The loss $\mathcal{L}(\delta, T)$ is the average negative log-likelihood (NLL) over the calibration set $\mathcal{D}_\text{calib}(x) = \{y_i\}_{i=1}^K$. Let $n_i$ denote the length of the sequence $y_i$. The NLL for a single answer sequence $y_i = (y_{i,1}, \dots, y_{i,n_i})$ is the sum of the NLLs for each of its $n_i$ generation steps:
$$ -\log p_\theta(y_i \mid x; \delta, T) = \sum_{j=1}^{n_i} -\log p_\theta(y_{i,j} \mid x, y_{i,<j}; \delta, T) $$
The total loss $\mathcal{L}$ is the average of this quantity over all $K$ sequences. Let $N = \sum_{i=1}^K n_i$ be the total number of generation steps across the entire calibration set. The loss can be written as an average over all these steps:
$$ \mathcal{L}(\delta, T) = \frac{1}{N} \sum_{i=1}^K \sum_{j=1}^{n_i} L_{i,j}(\delta, T) $$
where $L_{i,j}$ is the NLL for predicting token $y_{i,j}$ given the context $(x, y_{i,<j})$.
We then evaluate the gradient $\nabla \mathcal{L}(\delta, T) = \left[ \nabla_{\delta}\mathcal{L}, \frac{\partial \mathcal{L}}{\partial T} \right]$ at the initial point $(\mathbf{0}, 1)$.

\paragraph{Evaluating the Gradient Components at $(\delta,T)=(\mathbf{0},1)$.} \PYB{add $(\delta,T)=(\mathbf{0},1)$ to be clearer?}\YC{added.}
\subparagraph{Gradient with respect to $\delta$.}
At each generation step $(i, j)$, the shift $\delta$ is added to the hidden state $h_{i,j}$ before the final projection: $W_{\text{lm}} (h_{i,j} + \delta)$. The total gradient $\nabla_{\delta}\mathcal{L}$ is the average of the step-wise gradients. At $(\mathbf{0}, 1)$, this is:
\begin{equation}
\nabla_{\delta} \mathcal{L}(\mathbf{0}, 1) = \frac{1}{N} \sum_{i=1}^K \sum_{j=1}^{n_i} W_{\text{lm}}^\top \left( p_{i,j} - \mathbf{e}_{y_{i,j}} \right) = W_{\text{lm}}^\top \left( \bar{p}_{\theta} - \bar{p}_{\text{target}} \right)
\end{equation}
where $p_{i,j} = p_\theta(\cdot \mid x, y_{i,<j})$ is the base model's probability distribution for that step, $\mathbf{e}_{y_{i,j}}$ is the one-hot vector for the target token, $\bar{p}_{\theta} = \frac{1}{N}\sum_{i,j} p_{i,j}$ is the average predicted distribution, and $\bar{p}_{\text{target}} = \frac{1}{N}\sum_{i,j} \mathbf{e}_{y_{i,j}}$ is the average target distribution.

\subparagraph{Gradient with respect to $T$.}
Let $g_{i,j}$ be the vector of base model logits at step $(i, j)$. The step-wise loss is $L_{i,j}(T) = \log\left(\sum_{k=1}^{C} e^{g_{i,j,k}/T}\right) - \frac{g_{i,j,y_{i,j}}}{T}$. We now derive its partial derivative with respect to $T$:
\begin{align}
    \frac{\partial L_{i,j}(T)}{\partial T} &= \frac{\partial}{\partial T} \left[ \log\left(\sum_k e^{g_{i,j,k}/T}\right) \right] - \frac{\partial}{\partial T} \left[ \frac{g_{i,j,y_{i,j}}}{T} \right] \\
    &= \frac{1}{\sum_k e^{g_{i,j,k}/T}} \cdot \left( \sum_k e^{g_{i,j,k}/T} \cdot \frac{-g_{i,j,k}}{T^2} \right) + \frac{g_{i,j,y_{i,j}}}{T^2} \\
    &= \frac{g_{i,j,y_{i,j}}}{T^2} - \frac{1}{T^2} \sum_k \frac{e^{g_{i,j,k}/T}}{\sum_l e^{g_{i,j,l}/T}} \cdot g_{i,j,k} \\
    &= \frac{1}{T^2} \left( g_{i,j,y_{i,j}} - \mathbb{E}_{p_{i,j}(T)}[g_{i,j}] \right)
\end{align}
where $\mathbb{E}_{p_{i,j}(T)}[g_{i,j}]$ is the expected logit value under the softmax distribution with temperature $T$.

Evaluating at $T=1$ and averaging over all steps gives the total gradient for $T$:
\begin{equation}
    \frac{\partial \mathcal{L}}{\partial T} \Big|_{(\mathbf{0}, 1)} = \frac{1}{N} \sum_{i=1}^K \sum_{j=1}^{n_i} \left( g_{i,j,y_{i,j}} - \mathbb{E}_{p_{i,j}}[g_{i,j}] \right)
\end{equation}

\paragraph{The Joint Gradient is Non-Zero.}
The joint gradient is zero only if both of its components are zero.
\begin{enumerate}
    \item \textbf{$\delta$-gradient}: By our premise, $\bar{p}_{\theta} \neq \bar{p}_{\text{target}}$. The gradient $\nabla_{\delta} \mathcal{L}(\mathbf{0}, 1)$ is zero only if the non-zero vector $(\bar{p}_{\theta} - \bar{p}_{\text{target}})$ lies in the null space of $W_{\text{lm}}^\top$. This is equivalent to the error vector being orthogonal to the column space of $W_{\text{lm}}$. For large language models where $D \ll C$, this column space (of dimension at most $D$) is a very small subspace of $\mathbb{R}^C$. It is therefore highly improbable for a specific error vector, arising from model-data mismatch, to lie in the orthogonal complement of this subspace. Thus, we can assert with high confidence that $\nabla_{\delta} \mathcal{L}(\mathbf{0}, 1) \neq \mathbf{0}$.
    \item \textbf{$T$-gradient}: The $T$-gradient is zero only if $\frac{1}{N}\sum_{i,j} g_{i,j,y_{i,j}} = \frac{1}{N}\sum_{i,j} \mathbb{E}_{p_{i,j}}[g_{i,j}]$. This condition implies a perfect balance where the average logit of the ground-truth tokens equals the average expected logit over the vocabulary. An uncalibrated model typically exhibits systematic over-confidence (target logit is higher than the average, making the gradient negative) or under-confidence (target logit is lower, making the gradient positive). It is therefore highly unlikely for this gradient to be exactly zero unless the model is already well-calibrated in this specific sense.
\end{enumerate}
Given that the model is not perfectly calibrated, it is guaranteed that at least one of the gradient components is non-zero. Therefore, the joint gradient $\nabla \mathcal{L}(\mathbf{0}, 1)$ is non-zero.

\PYB{what is almost certain?} \YC{addressed.}

\paragraph{Existence of an Improving Solution.}
A non-zero gradient at the point $(\mathbf{0}, 1)$ implies the existence of a strict descent direction, $-\nabla \mathcal{L}(\mathbf{0}, 1)$. By Taylor's theorem for multivariate functions, for a small step $\alpha > 0$ in this direction, the new point $(\delta', T') = (\mathbf{0}, 1) - \alpha \nabla \mathcal{L}(\mathbf{0}, 1)$ will satisfy $\mathcal{L}(\delta', T') < \mathcal{L}(\mathbf{0}, 1)$. This proves the existence of an improving joint solution.

\end{proof}

% \subparagraph{Gradient with respect to $T$.}
% To derive the gradient with respect to temperature, we first consider the loss for a single, generic generation step. Let $g \in \mathbb{R}^C$ be the logit vector for this step, where $C$ is the vocabulary size. The corresponding NLL as a function of $T$ is:
% $$ L(T) = \log\left(\sum_{k=1}^{C} e^{g_k/T}\right) - \frac{g_{\text{target}}}{T} $$
% where $g_k$ is the $k$-th component of the logit vector and $g_{\text{target}}$ is the logit of the correct token. The partial derivative with respect to $T$ is:
% \begin{align}
%     \frac{\partial L(T)}{\partial T} &= \frac{1}{\sum_k e^{g_k/T}} \cdot \left( \sum_k e^{g_k/T} \cdot \frac{-g_k}{T^2} \right) + \frac{g_{\text{target}}}{T^2} \\
%     &= \frac{g_{\text{target}}}{T^2} - \frac{1}{T^2} \sum_k \frac{e^{g_k/T}}{\sum_l e^{g_l/T}} \cdot g_k \\
%     &= \frac{1}{T^2} \left( g_{\text{target}} - \mathbb{E}_{p(T)}[g] \right)
% \end{align}
% where $\mathbb{E}_{p(T)}[g]$ is the expected logit value under the softmax distribution with temperature $T$.

% Applying this result to each generation step $(i,j)$ of our problem (where the logits are $g_{i,j}$) and evaluating at $T=1$, the total gradient is the average over all steps:
% \begin{equation}
%     \frac{\partial \mathcal{L}}{\partial T} \Big|_{(\mathbf{0}, 1)} = \frac{1}{N} \sum_{i=1}^K \sum_{j=1}^{n_i} \left( g_{i,j,\text{target}} - \mathbb{E}_{p_{i,j}}[g_{i,j}] \right)
% \end{equation}

\subsection{Proof of Theorem ~\ref{theorem-1}}
\label{appendix:theorem-1}
\textbf{Theorem~\ref{theorem-1}} (Joint Calibration $(\delta, T)$ Improves Expected Reward from Best-of-N Sampling)
\begin{itshape}
Let $p_{\theta}(y \mid x; \delta, T)$ be the model's probability distribution over outputs $y \in \mathcal{Y}$, parameterized by a calibration vector $\delta$ and a temperature $T$. Let the base model be configured with parameters $(\mathbf{0}, T_{base})$ for some $T_{base} > 0$.
Let $R(x,y)$ be a reward function, and assume there exists a unique output $y^* \in \mathcal{Y}$ with a strictly maximum reward, i.e., $r^* = R(x,y^*) > \max_{y \neq y^*} R(x,y) = r_{\text{other\_max}}$.
We consider cases where joint calibration with parameters $(\delta^*, T^*)$ improves upon the base model by increasing the probability of the unique optimal output, i.e.,
$$p_{\theta}(y^* \mid x; \delta^*, T^*) > p_{\theta}(y^* \mid x; \mathbf{0}, T_{base})$$

Then, for any $n \geq 1$ within the remaining inference budget after calibration, the lower bound on the expected best-of-$N$ reward under the jointly calibrated model is strictly greater than that of the base model.
Specifically, let $R_{LB}(p) = r^* - (1-p)^n (r^* - r_{\text{other\_max}})$ be a valid lower bound for the expected best-of-$N$ reward, where $p$ is the probability of sampling $y^*$. The improvement in this lower bound, $\Delta_{R_{LB}}(x,n) = R_{LB}(p_{\theta}(y^* \mid x; \delta^*, T^*)) - R_{LB}(p_{\theta}(y^* \mid x; \mathbf{0}, T_{base}))$, is strictly positive.
\end{itshape}

%%% Previous therorem statement
% Let $p_{\theta}(y \mid x; \delta, T)$ be the model's probability distribution over outputs $y \in \mathcal{Y}$, parameterized by a calibration parameter $\delta$ and a temperature $T$.
% Let the base model be configured with parameters $(0, T)$.
% Let $R(x,y)$ be a reward function, and assume there exists a unique output $y^* \in \mathcal{Y}$ with a strictly maximum reward $r^* = R(x,y^*)$.
% Let $r_{\text{other\_max}} = \max_{y \neq y^*} R(x,y)$.

% We assume that joint calibration improves upon the base model by increasing the probability of the unique optimal output, i.e.,
% $$p_{\theta}(y^* \mid x; \delta^*, T^*) > p_{\theta}(y^* \mid x; 0, T)$$

% Then, for any number of samples $n \geq 1$, the lower bound on the expected best-of-N reward under the jointly calibrated model is strictly greater than the lower bound for the base model.

% Specifically, let $R_{LB}(p) = r^* - (1-p)^n (r^* - r_{\text{other\_max}})$ be a valid reward lower bound for the expected best-of-N reward, where $p$ is the probability of sampling $y^*$.
% The improvement in this lower bound $\Delta_{R_{LB}}(x,n) = L(p_{\theta}(y^* \mid x; \delta^*, T^*)) - L(p_{\theta}(y^* \mid x; 0, T))$ is strictly positive and is given by:
% $$
% \Delta_{R_{LB}}(x,n) = (r^* - r_{\text{other\_max}}) \left[ \left(1 - p_{\theta}(y^*|x; 0, T)\right)^n - \left(1 - p_{\theta}(y^*|x; \delta^*, T^*)\right)^n \right] > 0
% $$

% TODO: Reward Lower Bound L(p) -> R_L(p)

\begin{proof}

The proof proceeds in three steps.
First, we establish the expression for the expected best-of-N reward and derive its lower bound $R_{LB}(p)$.
Second, we prove that this lower bound $R_{LB}(p)$ is a strictly increasing function of $p$, the probability of sampling the optimal output $y^*$.
Finally, we use this monotonicity to prove the theorem's main claim.

\paragraph{Derivation of the Lower Bound $R_{LB}(p)$.}
\PYB{N or $N$? Be consistent} \YC{fixed}
Let $p = p_{\theta}(y^* \mid x)$ be the probability of sampling the unique optimal output $y^*$ in a single trial. The expected best-of-$N$ reward, $\mathbb{E}[\max_{i=1,...,n} R(x,y_i)]$, can be formulated by conditioning on whether $y^*$ is sampled at least once in $n$ trials.

The probability of sampling $y^*$ at least once is $1 - (1 - p)^n$. In this event, the maximum reward obtained is exactly $r^*$. The probability of never sampling $y^*$ in $n$ trials is $(1 - p)^n$. In this event, the expected maximum reward is $\mathbb{E}_{\text{other}} = \mathbb{E} [\max_{i=1,...,n} R(x,y_i) \mid \forall i, y_i \neq y^*]$.

The total expected reward is thus:
\begin{equation}
\begin{aligned}
    \mathbb{E} [\max_{i=1,...,n} R(x,y_i)] &= [1 - (1 - p)^n] r^* + (1 - p)^n \mathbb{E}_{\text{other}} \\
    &= r^* -(1 - p)^n (r^* - \mathbb{E}_{\text{other}})
\end{aligned}
\end{equation}

By definition, $\mathbb{E}_{\text{other}}$ is the expected maximum reward from a set of outputs where none is $y^*$. Therefore, this value cannot exceed the maximum possible reward in that set, $r_{\text{other\_max}}$. This gives the inequality $\mathbb{E}_{\text{other}} \leq r_{\text{other\_max}}$. Since $(r^* - x)$ is a decreasing function of $x$, this implies $(r^* - \mathbb{E}_{\text{other}}) \geq (r^* - r_{\text{other\_max}})$.

Substituting this into the expression for the expected reward yields a valid lower bound, which we denote $R_{LB}(p)$:
\begin{equation}
    \mathbb{E} [\max_{i=1,...,n} R(x,y_i)] \geq r^* - (1 - p)^n (r^* - r_{\text{other\_max}}) := R_{LB}(p)
\end{equation}

\paragraph{Prove the Monotonicity of $R_{LB}(p)$.}
We now show that $R_{LB}(p)$ is a strictly increasing function of $p$ for $p \in [0, 1)$. We take the derivative of $R_{LB}(p)$ with respect to $p$:
\begin{equation}
\begin{aligned}
    \frac{d R_{LB}}{dp} &= \frac{d}{dp} [r^* - (1 - p)^n(r^* - r_{\text{other\_max}})] \\
    &= -(-n (1-p)^{n-1})(r^* - r_{\text{other\_max}}) \\
    &= n (1-p)^{n-1} (r^* - r_{\text{other\_max}})
\end{aligned}
\end{equation}
By the theorem's assumptions, $n \geq 1$ and $r^* > r_{\text{other\_max}}$, which means $(r^* - r_{\text{other\_max}}) > 0$. For $p \in [0, 1)$, the term $(1-p)^{n-1}$ is also strictly positive. Therefore, $\frac{d R_{LB}}{dp} > 0$ for all $p \in [0, 1)$, which proves that $R_{LB}(p)$ is a strictly increasing function of $p$.

\paragraph{Conclusion.}
Let $p_{\text{cal}} = p_{\theta}(y^* \mid x; \delta^*, T^*)$ and $p_{\text{base}} = p_{\theta}(y^* \mid x; \mathbf{0}, T_{base})$. The theorem's central premise is $p_{\text{cal}} > p_{\text{base}}$.
Since $R_{LB}(p)$ is a strictly increasing function of $p$, the inequality $p_{\text{cal}} > p_{\text{base}}$ directly implies:
\begin{equation}
    R_{LB}(p_{\text{cal}}) > R_{LB}(p_{\text{base}})
\end{equation}
This proves that the lower bound on the expected reward is strictly greater for the calibrated model. The magnitude of this improvement, $\Delta_{R_{LB}}(x, n)$, is given by:
\begin{equation}
\begin{aligned}
    \Delta_{R_{LB}}(x, n) &= R_{LB}(p_{\text{cal}}) - R_{LB}(p_{\text{base}}) \\
    &= [r^* - (1 - p_{\text{cal}})^n (r^* - r_{\text{other\_max}})] \\
    & \enspace \quad - [r^* - (1 - p_{\text{base}})^n(r^* - r_{\text{other\_max}})] \\
    &= (r^* - r_{\text{other\_max}}) \left[ (1 - p_{\text{base}})^n - (1 - p_{\text{cal}})^n \right]
\end{aligned}
\end{equation}
Since $p_{\text{cal}} > p_{\text{base}}$, it follows that $(1 - p_{\text{cal}}) < (1 - p_{\text{base}})$. For $n \geq 1$, this implies $(1 - p_{\text{cal}})^n < (1 - p_{\text{base}})^n$. Thus, the term in the square brackets is strictly positive, and consequently $\Delta_{R_{LB}}(x, n) > 0$.

This completes the proof.

\end{proof}

\subsection{Proof of Corallary ~\ref{corollary-1}}
\label{appendix:corollary-1}
\textbf{Corollary~\ref{corollary-1}} (Sub-optimality of Exploitation Alone)
\begin{itshape}
The final candidate is selected by maximizing $R(x, y)$ over a set of candidates $\mathcal{Y}$. Since $\mathcal{Y}_{\text{exploit}}$ is a subset of the union $\mathcal{Y} = \mathcal{Y}_{\text{explore}} \cup \mathcal{Y}_{\text{exploit}}$, the strategy of only selecting from $\mathcal{Y}_{\text{exploit}}$ is sub-optimal compared to selecting from the union.
This is because the maximum reward achievable from the union is greater than or equal to the maximum reward achievable from the exploitation set alone.
$$
\max_{y \in \mathcal{Y}_{\text{explore}} \cup \mathcal{Y}_{\text{exploit}}} R(x, y) \geq \max_{y \in \mathcal{Y}_{\text{exploit}}} R(x, y)
$$

\end{itshape}

\begin{proof}

Let $R_{\text{explore}}^* = \max_{y \in \mathcal{Y}_{\text{explore}}} R(x, y)$ and $R_{\text{exploit}}^* = \max_{y \in \mathcal{Y}_{\text{exploit}}} R(x, y)$. The reward selected from the exploitation set is $R_{\text{exploit}}^*$, while the reward from the combined set is $R_{\text{final}} = \max_{y \in \mathcal{Y}_{\text{explore}} \cup \mathcal{Y}_{\text{exploit}}} R(x, y)$.

By the definition of the maximum function, the maximum of a set is greater than or equal to any of its elements. It follows that for any sampling outcome:
\begin{equation}
    R_{\text{final}} = \max(R_{\text{explore}}^*, R_{\text{exploit}}^*) \ge R_{\text{exploit}}^*
\end{equation}

This inequality holds universally for every possible generated set.
By the monotonicity of expectation, taking the expectation over all outcomes yields:
\begin{equation}
    \mathbb{E}[R_{\text{final}}] \ge \mathbb{E}[R_{\text{exploit}}^*]
\end{equation}
This demonstrates that retaining all $N_1 + N_2$ candidates yields an expected reward that is provably no worse than the reward obtained from the exploitation phase alone, and may in fact be better.
Therefore, only using the candidates from the exploitation phase is a suboptimal strategy.

\end{proof}

% \subsection{Proof of Lemma ~\ref{lemma-2}}
% \label{appendix:lemma-2}
% \input{docs/lemma_2}

% \subsection{Proof of Lemma ~\ref{lemma-3}}
% \label{appendix:lemma-3}
% \input{docs/lemma_3}

\section{Additional Results}
\subsection{Details of Reward-Guided Binary Search: Algorithm \& Motivation}
\label{appendix:toy-example}

\paragraph{Algorithm Description.}
Reward-guided binary search extends classic binary search by leveraging a reward model to guide the search process (see Algorithm~\ref{alg:reward-guided-binary-search} for pseudocode).
At each step, instead of simply splitting the interval in half, the algorithm first queries the reward at $n$ candidate points within the current interval.
The reward is designed as the inverse of the distance to the target, possibly perturbed by noise to reflect real-world model uncertainty.
Crucially, the algorithm selects the candidate point with the highest observed reward to refine the search interval, which similar to the strategy in test-time calibration, where high-reward completions are used as anchor points to guide subsequent exploration.
This approach fundamentally alters the sampling distribution: rather than always bisecting the interval, the search adaptively concentrates queries near regions with higher estimated reward, dynamically steering the search direction.
As a result, reward feedback not only accelerates convergence but also enables more efficient and adaptive exploration, especially when the reward model is reliable.

\paragraph{Reward Model and Noise.}
The reward function is defined as $r(x) = \frac{1}{|x - t|+1}$ where $t$ is the unknown target.
In practice, the reward model may be noisy due to imperfect estimation, so we add Gaussian noise: $r_{\text{obs}}(x) = r(x) + \epsilon, \quad \epsilon \sim \mathcal{N}(0, \sigma^2)$.
This noise models the fact that real-world reward models are not perfectly accurate and may deviate from the ground truth, making the search more challenging and realistic.

\paragraph{Motivation.}
Calibration (i.e., reward guiding) before each search step allows the algorithm to more efficiently narrow down the search space, especially when the reward model is reliable.
Importantly, calibration fundamentally alters the sampling distribution: instead of always splitting the interval at the midpoint, the algorithm adaptively selects query points based on reward feedback, concentrating samples near regions with higher estimated reward.
As shown in our experiments, increasing the number of calibration queries $n$ can dramatically reduce the number of search steps required, even under moderate noise.
This demonstrates the practical value of reward-guided search for tasks like TTS, where reward feedback reshapes the sampling distribution and accelerates convergence.

\begin{algorithm}[H]
\caption{Reward-Guided Binary Search with Calibration}
\label{alg:reward-guided-binary-search}
\KwIn{Search domain $[L, H]$, target $t$, calibration count $n$, reward noise $\sigma$}
\KwOut{Estimated target position}

\While{$L < H$}{
    \If{$n > 0$}{
        Select $n$ evenly spaced probe points $\{x_1, \ldots, x_n\}$ in $[L, H]$\;
        \ForEach{$x_i$}{
            Query reward: $r_i = \frac{1}{|x_i - t| + 1} + \epsilon_i$, $\epsilon_i \sim \mathcal{N}(0, \sigma^2)$\;
        }
        Let $x^* = \arg\max_{x_i} r_i$\;
        Estimate conservative bracket $[L', H']$ around $x^*$ using reward inversion and safety margin\;
        $L \gets \max(L, L')$\;
        $H \gets \min(H, H')$\;
        Set comparison point $x_c = \lfloor (L + H)/2 \rfloor$\;
    }
    \Else{
        Set comparison point $x_c = \lfloor (L + H)/2 \rfloor$\;
    }
    \If{$x_c < t$}{
        $L \gets x_c + 1$\;
    }
    \Else{
        $H \gets x_c$\;
    }
}
\Return{$L$}
\end{algorithm}

\subsection{Full Experimental Results}
\label{appendix:full-main-tables}

In this subsection, we provide the full experimental results for the MATH-500 benchmark, complementing the summary tables in the main text.

Table~\ref{table-math500-runtime} reports the total runtime (seconds) for each model under the same setup, comparing uncalibrated Best-of-$N$ and CarBoN methods across different rollout budget $N$.  
Importantly, CarBoN achieves higher accuracy at smaller rollout budgets while maintaining competitive total runtime.
For $N=64$, CarBoN surpasses the uncalibrated Best-of-$N$ at $N=256$ in accuracy for all models.  
Compared to the corresponding Best-of-$N$, the total runtime with CarBoN is lower for three of the four models, with the largest reduction for Qwen2.5-Math-1.5B-Instruct (27.19 sec vs.\ 46.22 sec), while for Llama-3.2-1B-Instruct the runtime is slightly higher (34.47 sec vs.\ 30.30 sec).

Table~\ref{table-math500} reports the corresponding accuracy results for both Vanilla and Weighted decoding across all four models and different $N$.  
For small $N$, Vanilla selection can occasionally achieve the highest accuracy, but for larger $N$ (128, 256), CarBoN consistently outperforms other methods, showing stable gains under both Vanilla and Weighted selection.  

Table~\ref{table-ablation} presents the ablation study for calibration parameters $(\delta, T)$ on two models (Llama-3.2-1B-Instruct and Qwen2.5-Math-1.5B-Instruct).  
Each experiment includes cases with only $\delta$, only $T$, or both combined (CarBoN), under both Vanilla and Weighted selection.  
While Vanilla selection occasionally achieves the highest accuracy for specific $N$ with a single parameter, Weighted decoding consistently performs best when combining $\delta$ and $T$, highlighting the complementary benefits of the two calibration parameters.  
All values are reported as accuracy (\%).  

\begin{table}[t]
\centering
\caption{
\textbf{Total runtime (seconds) of four models on the MATH-500 benchmark, comparing Weighted Best-of-$N$ methods before and after calibration.}  
CarBoN achieves higher accuracy at lower rollout budgets, while maintaining comparable or faster total runtime relative to the corresponding uncalibrated Best-of-$N$: for example, $N=64$ with CarBoN outperforms the uncalibrated Best-of-$N$ at $N=256$ in accuracy.
}
\label{table-math500-runtime}
\vspace{-0.8em}
\resizebox{0.85\textwidth}{!}{
\begin{tabular}{ccrrrrrr}
\toprule
\multirow{2}{*}{Model} & \multirow{2}{*}{Method} & \multicolumn{6}{c}{N} \\ 
\cmidrule(lr){3-8} 
 & & \multicolumn{1}{c}{8} & \multicolumn{1}{c}{16} & \multicolumn{1}{c}{32} & \multicolumn{1}{c}{64} & \multicolumn{1}{c}{128} & \multicolumn{1}{c}{256} \\
\midrule
\multirow{2}{*}{Llama-3.2-1B-Instruct}      
& Best-of-$N$ & 4.26 & 5.46 & 7.20 & 10.76 & 17.66 & 30.30 \\
& CarBoN    & 6.27 & 11.09 & 16.82 & 34.47 & 47.98 & 82.15 \\
\midrule
\multirow{2}{*}{Llama-3.2-3B-Instruct}      
& Best-of-$N$ & 5.06 & 7.70 & 9.68 & 14.08 & 19.14 & 32.48 \\
& CarBoN    & 7.36 & 13.33 & 19.29 & 29.74 & 49.27 & 88.92 \\
\midrule
\multirow{2}{*}{Qwen2.5-1.5B-Instruct}      
& Best-of-$N$ & 5.44 & 10.92 & 15.10 & 17.00 & 25.72 & 44.14 \\
& CarBoN    & 8.25 & 14.00 & 20.59 & 31.54 & 55.23 & 99.59 \\
\midrule
\multirow{2}{*}{Qwen2.5-Math-1.5B-Instruct} 
& Best-of-$N$ & 7.80 & 9.54 & 15.32 & 16.22 & 25.62 & 46.22 \\
& CarBoN    & 9.39 & 12.49 & 20.93 & 27.19 & 60.56 & 113.05 \\
\bottomrule
\end{tabular}
}
\end{table}
\begin{table}[ht]
\centering
\caption{
\textbf{Accuracy (\%) of four models on the MATH-500 benchmark, comparing Vanilla and Weighted Best-of-$N$ methods before and after calibration.}
CarBoN enables further improvements beyond the plateau of standard Best-of-$N$, with calibrated accuracy at $N=64$ exceeding the uncalibrated results at $N=256$, corresponding to up to $4\times$ less rollout budgets.
Bold numbers indicate the better accuracy \emph{within the same method type} (Vanilla vs. Vanilla, Weighted vs. Weighted) for each $N$. 
}
\vspace{-0.8em}
\resizebox{0.85\textwidth}{!}{
\begin{tabular}{cclrrrrrr}
\toprule
% \multicolumn{9}{c}{MATH-500}                                                                                                                                                                                                                          \\
% \midrule
\multirow{2}{*}{Model}                      & \multicolumn{2}{c}{\multirow{2}{*}{Method}}      & \multicolumn{6}{c}{N}                                                                                                                                \\ \cmidrule(lr){4-9} 
                                            & \multicolumn{2}{c}{}                             & \multicolumn{1}{c}{8} & \multicolumn{1}{c}{16} & \multicolumn{1}{c}{32} & \multicolumn{1}{c}{64} & \multicolumn{1}{c}{128} & \multicolumn{1}{c}{256} \\
\midrule
\multirow{4}{*}{Llama-3.2-1B-Instruct}      & \multirow{2}{*}{Best-of-$N$}            & Vanilla  & \textbf{44.2}       & 45.8                 & 47.6                 & \textbf{49.4}        & 49.8                  & 50.0                  \\
                                            &                                       & Weighted & 42.0                & 44.6                 & 47.8                 & 48.6                 & 50.6                  & 50.8                  \\ \cmidrule(lr){2-9} 
                                            & \multirow{2}{*}{CarBoN}               & Vanilla  & 43.8                & \textbf{47.4}       & \textbf{48.0}        & 49.0                 & \textbf{50.0}         & \textbf{50.6}         \\
                                            &                                       & Weighted & \textbf{43.0}       & \textbf{45.6}        & \textbf{48.4}        & \textbf{51.0}        & \textbf{51.8}         & \textbf{51.8}         \\
\midrule
\multirow{4}{*}{Llama-3.2-3B-Instruct}      & \multirow{2}{*}{Best-of-$N$}            & Vanilla  & 56.8                & \textbf{59.0}        & 60.0                 & 60.2                 & 60.4                  & 60.6                  \\
                                            &                                       & Weighted & 56.8                & 58.2                 & 59.6                 & 61.6                 & 61.8                  & 62.2                  \\ \cmidrule(lr){2-9} 
                                            & \multirow{2}{*}{CarBoN}               & Vanilla  & \textbf{57.0}       & 58.8                 & \textbf{61.0}        & \textbf{61.6}        & \textbf{61.6}         & \textbf{61.8}         \\
                                            &                                       & Weighted & \textbf{57.6}       & \textbf{59.0}        & \textbf{60.8}        & \textbf{62.2}        & \textbf{63.2}         & \textbf{63.4}         \\
\midrule
\multirow{4}{*}{Qwen2.5-1.5B-Instruct}      & \multirow{2}{*}{Best-of-$N$}            & Vanilla  & \textbf{53.6}       & \textbf{54.4}        & \textbf{55.2}        & 55.8                 & 56.2                  & 56.0                  \\
                                            &                                       & Weighted & \textbf{56.4}       & 57.6                 & 61.4                 & 62.0                 & 62.6                  & 62.2                  \\ \cmidrule(lr){2-9} 
                                            & \multirow{2}{*}{CarBoN}               & Vanilla  & 50.8                & 53.8                 & 53.8                 & \textbf{56.8}        & \textbf{56.6}         & \textbf{59.6}         \\
                                            &                                       & Weighted & 55.0                & \textbf{60.0}        & \textbf{61.8}        & \textbf{62.4}        & \textbf{64.0}         & \textbf{64.4}         \\
\midrule
\multirow{4}{*}{Qwen2.5-Math-1.5B-Instruct} & \multirow{2}{*}{Best-of-$N$}            & Vanilla  & \textbf{71.0}       & 70.8                 & 71.0                 & 70.8                 & 70.8                  & 70.8                  \\
                                            &                                       & Weighted & 73.6                & 75.4                 & 76.4                 & 75.6                 & 76.4                  & 76.8                  \\ \cmidrule(lr){2-9} 
                                            & \multirow{2}{*}{CarBoN}               & Vanilla  & 70.4                & \textbf{71.2}        & \textbf{73.0}        & \textbf{72.6}        & \textbf{73.4}         & \textbf{73.4}         \\
                                            &                                       & Weighted & \textbf{74.2}       & \textbf{76.0}        & 76.4                 & \textbf{77.2}        & \textbf{77.2}         & \textbf{77.8}         \\

\bottomrule
\end{tabular}
}
\label{table-math500}
\end{table}
\begin{table}[ht]
\centering
\caption{
\textbf{Ablation study on calibration parameters $(\delta, T)$ and their combination (CarBoN) for Best-of-$N$ search on MATH-500.}
We compare vanilla Best-of-$N$, applying a shift ($\delta$), a temperature scaling ($T$), and their joint calibration (CarBoN), under both vanilla and weighted selection strategies.
All values report accuracy (\%).
Results show that CarBoN consistently improves accuracy across different $N$, especially with the weighted variant, highlighting the complementary benefits of $\delta$ and $T$.
Bold numbers indicate the better accuracy \emph{within the same method type} (Vanilla vs. Vanilla, Weighted vs. Weighted) for each $N$.
}
\vspace{-0.8em}
\resizebox{0.85\textwidth}{!}{

\begin{tabular}{cclrrrrrr}
\toprule
\multirow{2}{*}{Model}                      & \multicolumn{2}{c}{\multirow{2}{*}{Method}}                   & \multicolumn{6}{c}{N}                                                                                                                                \\ \cmidrule(lr){4-9} 
                                            & \multicolumn{2}{c}{}                                          & \multicolumn{1}{c}{8} & \multicolumn{1}{c}{16} & \multicolumn{1}{c}{32} & \multicolumn{1}{c}{64} & \multicolumn{1}{c}{128} & \multicolumn{1}{c}{256} \\
\midrule
\multirow{8}{*}{Llama-3.2-1B-Instruct}      & \multirow{2}{*}{Best-of-$N$}             & Vanilla              & \textbf{44.2}       & 45.8                 & 47.6                 & \textbf{49.4}        & 49.8                  & 50.0                  \\
                                            &                                        & Weighted             & 42.0                & 44.6                 & 47.8                 & 48.6                 & 50.6                  & 50.8                  \\ \cmidrule(lr){2-9} 
                                            & \multirow{2}{*}{Best-of-$N$ w/ $\delta$} & Vanilla              & 42.8                & 46.6                 & \textbf{48.8}        & 48.2                 & 49.6                  & 50.0                  \\
                                            &                                        & Weighted             & 41.8                & 43.8                 & 48.2                 & 49.0                 & 51.0                  & 51.2                  \\ \cmidrule(lr){2-9}  
                                            & \multirow{2}{*}{Best-of-$N$ w/ $T$}      & Vanilla              & 42.4                & \textbf{47.6}        & 46.8                 & 48.4                 & \textbf{50.4}         & 50.4                  \\
                                            &                                        & Weighted             & 42.0                & 45.0                 & 47.0                 & 49.6                 & 49.8                  & 50.6                  \\ \cmidrule(lr){2-9}  
                                            & \multirow{2}{*}{CarBoN}                & Vanilla              & 43.8                & 47.4                 & 48.0                 & 49.0                 & 50.0                  & \textbf{50.6}         \\
                                            &                                        & Weighted             & \textbf{43.0}       & \textbf{45.6}        & \textbf{48.4}        & \textbf{51.0}        & \textbf{51.8}         & \textbf{51.8}         \\
\midrule
\multirow{8}{*}{Qwen2.5-Math-1.5B-Instruct} & \multirow{2}{*}{Best-of-$N$}             & Vanilla            & \textbf{71.0}       & 70.8                 & 71.0                 & 70.8                 & 70.8                  & 70.8                  \\
                                            &                                        & Weighted             & 73.6                & 75.4                 & \textbf{76.4}        & 75.6                 & 76.4                  & 76.8                  \\ \cmidrule(lr){2-9} 
                                            & \multirow{2}{*}{Best-of-$N$ w/ $\delta$} & Vanilla            & \textbf{71.0}       & 71.0                 & 71.2                 & 71.0                 & 70.8                  & 70.8                  \\
                                            &                                        & Weighted             & \textbf{74.2}       & 74.8                 & 75.6                 & 77.0                 & 76.6                  & 77.0                  \\ \cmidrule(lr){2-9} 
                                            & \multirow{2}{*}{Best-of-$N$ w/ $T$}      & Vanilla            & 70.0                & \textbf{71.2}        & 71.0                 & 70.8                 & 70.8                  & 70.8                  \\
                                            &                                        & Weighted             & 73.2                & 75.2                 & 76.0                 & 76.4                 & 76.0                  & 76.6                  \\ \cmidrule(lr){2-9} 
                                            & \multirow{2}{*}{CarBoN}                & Vanilla              & 70.4                & \textbf{71.2}        & \textbf{73.0}        & \textbf{72.6}        & \textbf{73.4}         & \textbf{73.4}         \\
                                            &                                        & Weighted             & \textbf{74.2}       & \textbf{76.0}        & \textbf{76.4}        & \textbf{77.2}        & \textbf{77.2}         & \textbf{77.8}         \\
\bottomrule
\end{tabular}

}
\label{table-ablation}
\end{table}

\subsection{Supplementary Results on Larger Models (Pass@1)}
\label{appendix:bigger_llm}
For reference, we additionally report results on several larger closed-source and open-source models, evaluated under the same setup as in the main experiments (identical system prompt as shown in Appendix~\ref{appendix:system_prompt} and sampling temperature $T=0.8$).
These results provide additional context, illustrating that test-time scaling with smaller models can approach the performance level of substantially larger models.
Table~\ref{tab:bigger_llm} reports single-sample (@1) accuracies for the selected models.

\begin{table}[t]
\centering
\caption{\textbf{Pass@1 accuracy on larger closed-source and open-source models.} Evaluated under the same setup as in the main experiments (system prompt and $T=0.8$), these results serve as reference points for comparing test-time scaling with smaller models.}
\label{tab:bigger_llm}
\resizebox{0.8\textwidth}{!}{

\begin{tabular}{l l cc}
\toprule
\textbf{Type} & \textbf{Model} & \textbf{MATH-500} & \textbf{AIME-2024} \\
\midrule
\multirow{3}{*}{Closed Source LLMs}
& gpt-5-nano & -- & 11/30 \\
& o1-preview & 87.0\% & 12/30 \\
& gpt-4o & 77.0\% & 4/30 \\
\midrule
\multirow{3}{*}{Open Source LLMs}
& Qwen2.5-Math-7B-Instruct & 73.2\% & 4/30 \\
& Qwen2.5-7B-Instruct & 61.4\% & 3/30 \\
& LLaMA-3.1-8B-Instruct & 41.4\% & 3/30 \\
\bottomrule
\end{tabular}

}
\end{table}

\subsection{Temperature Scaling with Sample Size $N$}
\label{appendix:temp_vs_n}

\begin{figure*}[t]
    \centering
    \includegraphics[width=0.9\linewidth]{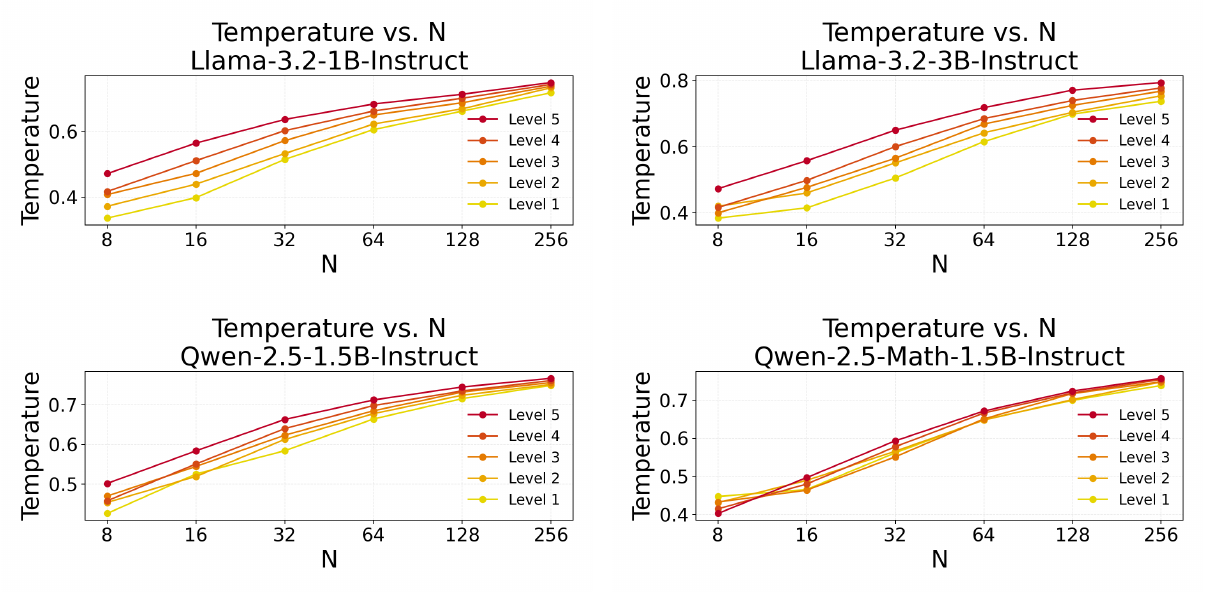}
    \caption{
    \textbf{Learned calibration temperatures across different rollout budget $N$ and problem difficulty levels.}
    Both larger $N$ and higher difficulty consistently lead to higher $T$, reflecting the increased diversity of top-$k$ completions.
    This adaptive scaling complements our earlier grid-search results in Appendix~\ref{appendix:llama-temp}, where small $N$ favored lower temperatures but larger $N$ required higher temperatures to fully leverage the broader exploration.
    }
    \label{fig:temp_vs_n}
\end{figure*}

We further investigate how the learned calibration temperature varies with the sample size $N$. 
Figure~\ref{fig:temp_vs_n} plots the temperature across five difficulty levels as $N$ increases. 
Two consistent trends emerge:
(i) more difficult problems require higher temperatures, as discussed in the main text, and
(ii) larger $N$ also leads to higher optimal temperatures. 
The latter reflects that with more samples, a higher temperature is necessary to encourage sufficient diversity and thereby better utilize the expanded inference budget. 
Otherwise, generating many samples under a low temperature yields near-identical outputs, effectively wasting the additional budget. 
A simple example is buying 100 lottery tickets with the same number (low $T$), where even with more tickets the outcome remains largely unchanged.
With diverse numbers (high $T$), additional tickets meaningfully increase the chance of winning.

This finding aligns with our earlier grid-search experiments (Appendix~\ref{appendix:llama-temp}).
For small $N$, lower temperatures tend to reach peak accuracy earlier, while higher temperatures show little improvement initially but yield greater gains for larger $N$.
This indicates that temperature affects the growth rate of accuracy, and that a fixed temperature may either converge early (low $T$) or show delayed benefits (high $T$), highlighting the need to adapt $T$ to $N$.
Importantly, adapting $T$ to $N$ also explains why baseline methods with a fixed temperature tend to converge sooner than CarBoN, while CarBoN continues to improve as $N$ increases.
Together, these results highlight that temperature is not a fixed hyperparameter, but should adapt naturally to both task difficulty and inference-time compute.

\subsection{Correlation between Problem Difficulty and Calibration Statistics}
\label{appendix:cor_level_temp}

To further examine how calibration relates to problem difficulty, we computed 
Spearman rank correlations ($\rho$) in our best-of-$N$ study with $N=256$ setting, where $N_1=128$ samples are generated for exploration (select top-$k$ score completions to construct the calibration dataset), and $N_2=128$ samples are used in the second phase for exploitation.

Specifically, we considered two quantities:  
(i) the entropy of the calibration dataset constructed from top-$k$ completions in the exploration phase ($N_1$), and  
(ii) the learned temperature estimated from the calibration dataset and applied in the exploitation phase ($N_2$). 

As shown in Table~\ref{tab:spearman}, the learned temperature exhibits an almost perfect 
correlation with problem difficulty across all four models ($\rho \approx 0.99999$), $p < 10^{-5}$). 
For the calibration dataset entropy, three models exhibit near-perfect correlations, while Qwen2.5-1.5B-Instruct shows a slightly weaker yet still strong correlation
(\(\rho \approx 0.9\), $p \approx 0.037$).

\begin{table}[t]
\centering
\caption{Spearman rank correlations ($\rho$) in the best-of-$N$ setting (rollout budget $N=256$), between problem difficulty and 
calibration dataset entropy from top-$k$ completions ($N_1=128$, $k=32$ for calibration dataset), and 
learned calibration temperature ($N_2=128$).}
\label{tab:spearman}
\resizebox{0.65\textwidth}{!}{
\begin{tabular}{lcc}
\toprule
\textbf{Model} & \textbf{Entropy} & \textbf{Temperature} \\
\midrule
Llama-3.2-1B-Instruct   & 0.999 & 0.999 \\
Llama-3.2-3B-Instruct   & 0.999 & 0.999 \\
Qwen2.5-1.5B-Instruct   & 0.899 & 0.999 \\
Qwen2.5-Math-1.5B-Instruct & 0.999 & 0.999 \\
\bottomrule
\end{tabular}
}
\end{table}

\subsection{Top-$k$ Token Overlap Metrics for High-Scoring Answers}
\label{appendix:delta_token_metric}

We formalize four token-level overlap metrics that quantify how closely a group of generated answers (either after calibration with $\delta$ or an uncalibrated set) aligns lexically with the vocabulary used by the top-$k$ highest-scoring answers for each problem.
All metrics are computed independently per problem and then macro-averaged across the full dataset (i.e., each problem contributes equally regardless of length).

Let $\text{Target}$ denote the de-duplicated set of token IDs appearing in the union of the top-$k$ high-scoring answers for a given problem with special tokens removed.
Let $X$ be the comparison token set built from either (i) calibrated generations after applying $delta$ ("calibration w/ $\delta$") or (ii) the uncalibrated ("No Calibration").
Define: \[ I = |\text{Target} \cap X|, \quad U = |\text{Target} \cup X|, \quad n_{\text{Target}} = |\text{Target}|, \quad n_X = |X|. \]

We report:
\begin{itemize}
    \item Jaccard similarity: $J(\text{Target}, X) = \frac{|\text{Target} \cap X|}{|\text{Target} \cup X|} = \frac{I}{U}.$
    \item Dice (Sørensen–Dice) coefficient: $D(\text{Target}, X) = \frac{2|\text{Target} \cap X|}{|\text{Target}| + |X|} = \frac{2I}{n_{\text{Target}} + n_X} = \frac{2J}{1+J}.$
    \item Recall (coverage of high-scoring tokens): $\text{Recall}(\text{Target} \rightarrow X) = \frac{|\text{Target} \cap X|}{|\text{Target}|} = \frac{I}{n_{\text{Target}}}.$
    \item Precision (specificity toward high-scoring tokens): $\text{Precision}(\text{Target} \leftarrow X) = \frac{|\text{Target} \cap X|}{|X|} = \frac{I}{n_X}.$

\end{itemize}

\paragraph{Interpretation.}
Jaccard and Dice provide set-level similarity that penalizes both omissions (missing reference tokens) and additions (extra tokens outside the reference).
Recall measures how completely the high-quality lexical signal is covered by the generated group, while Precision measures how selectively the group reuses only that high-quality signal (penalizing off-pattern or noisy additions).
A calibration that increases Jaccard/Dice and Precision while maintaining high Recall indicates convergence toward the lexical core of high-scoring answers without excessive loss of useful diversity.

All four metrics are first computed per problem and then averaged uniformly across all problems (macro average). This prevents problems with larger token sets from dominating the aggregate.

\section{Experiment Details}
% \subsection{Computing Resources}
% We conducted our experiments on two types of nodes: (i) two nodes with 4 × NVIDIA H100 (80GB) GPUs, 96 CPU cores, and 1 TB of RAM; and (ii) four nodes with 8 × NVIDIA RTX 3090 (24GB) GPUs, up to 44 CPU cores, and 768 GB of RAM. All nodes ran Debian 12. This setup provided sufficient computational resources for efficient data processing and analysis.
\subsection{Computational Environment}
Experiments were conducted on two types of nodes: (i) two nodes with 4 × NVIDIA H100 (80GB) GPUs, 96 CPU cores, and 1 TB RAM; and (ii) two nodes with 8 × NVIDIA RTX 3090 (24GB) GPUs, up to 44 CPU cores, and 768 GB RAM. All experiments used Python 3.11.11, PyTorch 2.4.0, vLLM 0.6.3, and CUDA 12.9.

\subsection{Temperature Grid Search.}
\label{appendix:llama-temp}
\begin{figure*}[t!bp]
    \centering
    \includegraphics[width=\linewidth]{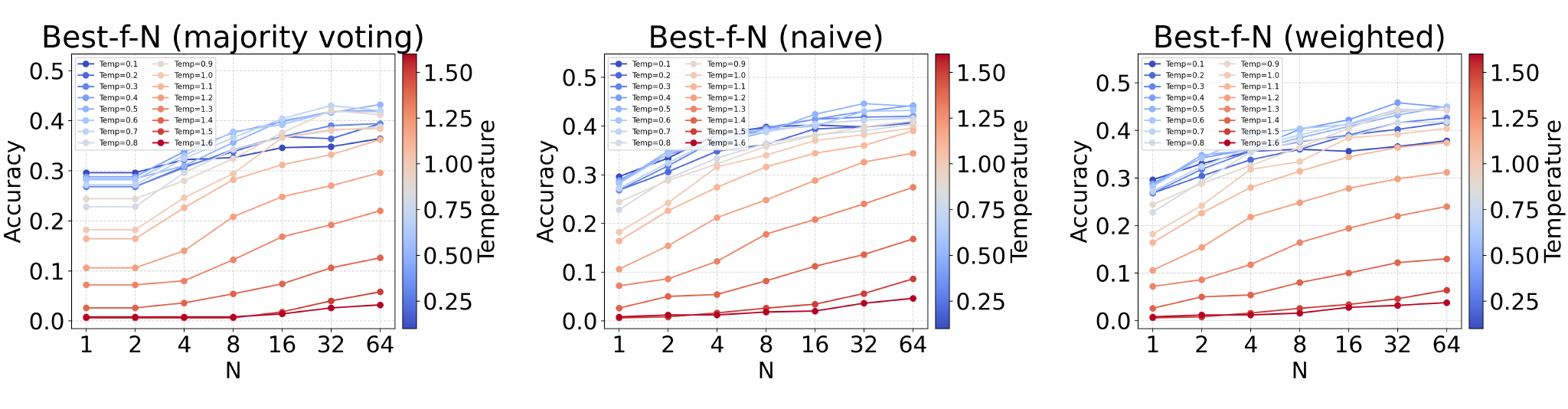}
    \caption{Results of Llama-3.2-1B-Instruct on MATH-500 with different temperatures $T$ for best-of-$N$ inference. From left to right: majority voting, naive, and weighted selection. Blue curves indicate lower temperatures, red curves indicate higher temperatures. Lower temperatures improve accuracy, but overly low temperatures limit diversity and the benefit of increasing $N$.}
    \label{fig:llama-temp}
\end{figure*}

Figure~\ref{fig:llama-temp} shows the results of Llama-3.2-1B-Instruct on the MATH-500 dataset using different temperatures ($T \in [0.1, 1.6]$ with step size $0.1$)) for best-of-$N$ inference, with majority voting (left), naive (middle), and weighted (right) selection strategies.
Across all settings of $N=1,2,4,\dots,64$, lower temperatures (blue curves) consistently yield higher accuracy by focusing generation on high-quality answers. However, when the temperature is too low, the improvement with increasing $N$ becomes marginal, especially for majority voting and weighted selection, indicating that excessive concentration limits diversity and exploration.
Conversely, higher temperatures (red curves) increase diversity but reduce accuracy for mathematical problems; at extremely high temperatures (e.g., $T=1.6$), the model struggles to solve the tasks and accuracy drops sharply, highlighting the importance of careful temperature tuning.
Previous findings \citep{beeching2024scalingtesttimecompute} show that sampling with $T=1.0$ sometimes leads the model to unexpectedly generate Chinese characters mid-solution and hurts performance. Considering both accuracy and exploration, and following previous work \citep{snell2024scaling}, we adopt $T=0.8$ as the baseline temperature for all experiments.

\subsection{Calibration Training Details.}
\label{appendix:calibration-training}
Calibration parameters $(\delta, T)$ are optimized using AdamW in a full-batch setting for 100 epochs per problem.
The learning rate is $0.001$ with a constant schedule, and a weight decay of $10^{-2}$ is applied only to $\delta$.
The parameters are initialized as $\delta=0$ and $T=0.8$.
The loss is the negative log-likelihood over the top-$k$ candidates.

For each evaluation budget $N$, we split it evenly into $N_1 = N_2 = N/2$. In the first stage, $N_1$ completions are generated at $T=0.8$, and the top-$k$ highest-scoring completions ($k = N_1/4$) form the calibration dataset.
The calibration parameters $(\delta, T)$ are then trained directly on cached logits.
In the second stage, the remaining $N_2$ completions are generated using the learned $(\delta, T)$.

This procedure ensures lightweight test-time calibration, requiring no additional model forward passes beyond the initial generation.

\subsection{System Prompt}
\label{appendix:system_prompt}
For all test-time scaling experiments, we follow previous work \cite{snell2024scaling} and adopt the same system prompt across all models to ensure consistency and comparability of results. This allows us to isolate the effects of the scaling methods without introducing variability from different prompts, as detailed in Table \ref{tab:system_prompt}.

\begin{table}[ht]
% \small
\centering
\caption{System Prompt for all experiments}
\label{tab:system_prompt}

\begin{tabular}{@{}p{1\columnwidth}@{}}
\toprule
\textbf{System Prompt}\\
\midrule

Solve the following math problem efficiently and clearly: \\
\\
- For simple problems (2 steps or fewer): \\
Provide a concise solution with minimal explanation.\\
- For complex problems (3 steps or more): \\
Use this step-by-step format: \\
\#\# Step 1: [Concise description] \newline
[Brief explanation and calculations] \\
\#\# Step 2: [Concise description] \newline
[Brief explanation and calculations] \\
...\\
Regardless of the approach, always conclude with:\\
Therefore, the final answer is: $\text{\$\textbackslash\textbackslash boxed\{answer\}\$}$. I hope it is correct.\\
Where [answer] is just the final number or expression that solves the problem. \\

\bottomrule
\end{tabular}
\end{table}

\end{document}